\documentclass[preprint,12pt]{elsarticle}

\usepackage{booktabs}
\usepackage{pifont}%
\usepackage{rotating}
\usepackage{multirow}
\usepackage{color}
\usepackage{graphicx}
\usepackage{hyperref}
\usepackage{amsfonts}
\usepackage{amsmath}
\usepackage{amssymb}
\usepackage[noend]{algpseudocode}
\usepackage{varwidth}
\usepackage{algorithmicx}
\usepackage{nicefrac}
\usepackage{algorithm} %
\usepackage{subcaption}
\usepackage{wrapfig}
\usepackage{enumitem}
\usepackage{xspace}
\usepackage[normalem]{ulem}
\usepackage{listings}
\usepackage{placeins}
\usepackage{glossaries-extra}

\usepackage{tikz}
\usetikzlibrary{positioning, shapes}

\newcommand{\DefMacro}[2]{\expandafter\newcommand\csname rmk-#1\endcsname{#2}}
\newcommand{\UseMacro}[1]{\csname rmk-#1\endcsname}

\newcommand{\CodeIn}[1]{\begin{small}\texttt{#1}\end{small}}
\newcommand{\XComment}[1]{}
\newcommand{\Space}[1]{}

\newcommand{\TNC}[2]{\multicolumn{#1}{c}{#2}}

\newcommand{\HighlightCell}[1]{\textbf{#1}}

\definecolor{gray}{RGB}{211,211,211}
\newcommand{\jbasicstyle}{\small\sffamily}

\newcommand{\jnumberstyle}{\scriptsize}

\usepackage[disable,textwidth=60pt,textsize=scriptsize]{todonotes} %
\LetLtxMacro{\todom}{\todo}
\renewcommand{\todo}[1]{\todom[inline]{#1}}

\lstdefinelanguage{pseudo}
{ morekeywords={for, in, break, continue, try, except, not,
  if,else,return,map,fieldElement_array_array40,fieldElement_array40},
  keywordstyle=\bfseries, lineskip=-0.1em, numbers=left,
  numberstyle=\jnumberstyle, numbersep=4pt, basicstyle=\jbasicstyle,
  breaklines=true, breakautoindent=true, tabsize=2,
  columns=fullflexible, morecomment=*[l][\textsl]{//},
  mathescape=true, }
\definecolor{javared}{rgb}{0.6,0,0} %
\definecolor{javagreen}{rgb}{0.25,0.5,0.35} %
\definecolor{javapurple}{rgb}{0.5,0,0.35} %
\definecolor{javadocblue}{rgb}{0.25,0.35,0.75} %
 
\setabbreviationstyle[acronym]{long-short}

\newacronym{drl}{DRL}{Deep Reinforcement Learning}
\newcommand{\DRL}{\gls*{drl}\xspace}

\newacronym{mdp}{MDP}{Markov Decision Process}
\newcommand{\MDP}{\gls*{mdp}\xspace}

\newacronym{moe}{\textsc{MoE}}{Mixture of Experts}
\newcommand{\MOE}{\gls*{moe}\xspace}

\newacronym{em}{\textsc{EM}}{Expectation-Maximization}
\newcommand{\EM}{\gls*{em}\xspace}

\newacronym{dt}{DT}{decision tree}
\newcommand{\DT}{\gls*{dt}\xspace}
\newcommand{\DTs}{\gls*{dt}s\xspace}

\newacronym{dsl}{DSL}{Domain-Specific Language}

\newacronym{dqn}{DQN}{deep Q-network}
\newcommand{\DQN}{\gls*{dqn}\xspace}

\DeclareMathOperator*{\argmax}{arg\,max}

\newcommand{\Section}[1]{\section{#1}}
\newcommand{\BeginTable}{\begin{table}[!t] \small}
\newcommand{\BeginTableTwoColumn}{\begin{table*}[!t] \small}
\newcommand{\EndTable}{\end{table}}
\newcommand{\EndTableTwoColumn}{\end{table*}}
\newcommand{\BeginEquation}{\begin{equation}}

\renewcommand{\vec}[1]{\mathbf{#1}}

\newcommand{\Cartpole}{CartPole\xspace}
\newcommand{\Pong}{Pong\xspace}
\newcommand{\Acrobot}{Acrobot\xspace}
\newcommand{\Mountaincar}{Mountaincar\xspace}
\newcommand{\Lunarlander}{Lunarlander\xspace}
\newcommand{\Pendulum}{Pendulum\xspace}

\newcommand{\OpenAI}{OpenAI\xspace}
\newcommand{\OpenAIGym}{OpenAI Gym\xspace}

\newcommand{\Tool}{\textsc{Mo\"ET}\xspace}
\newcommand{\MOET}{\Tool}
\newcommand{\MOEHard}{\textsc{Mo\"ET}\textsubscript{h}\xspace}
\newcommand{\MOETHard}{\MOEHard}
\newcommand{\MOETH}{\MOEHard}

\newcommand{\fig}{Figure\xspace}

\newcommand{\Equat}{Eq.\xspace}

\newcommand{\MOETbl}{\Tool\xspace}

\newcommand{\Viper}{Viper\xspace}
\newcommand{\Dagger}{\textsc{DAgger}\xspace}

\newcommand{\zt}{Z3\xspace}

\newcommand{\Qvalue}{Q-value\xspace}

\algnewcommand\algorithmicforeach{\textbf{for each}}
\algdef{S}[FOR]{ForEach}[1]{\algorithmicforeach\ #1\ \algorithmicdo}

\DefMacro{cartpole_DRL_reward}{200.00}
\DefMacro{pong_DRL_reward}{21.00}
\DefMacro{acrobot_DRL_reward}{-68.60}
\DefMacro{mountaincar_DRL_reward}{-105.27}

\usepackage[export]{adjustbox}

\newcommand{\CartpoleSaturationDepth}{$8$\xspace}
\newcommand{\PongSaturationDepth}{$30$\xspace}
\newcommand{\AcrobotSaturationDepth}{$15$\xspace}
\newcommand{\MountaincarSaturationDepth}{$12$\xspace}
\newcommand{\LunarlanderSaturationDepth}{$20$\xspace}
\newcommand{\PendulumSaturationDepth}{$20$\xspace}

\newlength{\textfloatsepsave}
\journal{Neural Networks}

\begin{document}

\begin{frontmatter}

\title{Mo\"ET: Mixture of Expert Trees and its Application to Verifiable Reinforcement Learning}

\author[TEX]{Marko Vasic}\ead{vasic@utexas.edu}
\author[SIN]{Andrija Petrovic}
\author[GOOG]{Kaiyuan Wang}
\author[MATF]{Mladen Nikolic}
\author[GBRAIN]{Rishabh Singh}
\author[TEX]{Sarfraz Khurshid}

\address[TEX]{The University of Texas at Austin, USA}
\address[SIN]{Singidunum University, Serbia}
\address[GOOG]{Google, USA}
\address[MATF]{University of Belgrade, Serbia}
\address[GBRAIN]{Google Brain, USA}

\begin{abstract}
Rapid advancements in deep learning have led to many recent breakthroughs.
While deep learning models achieve superior performance, often statistically better than humans, their adoption into safety-critical settings, such as healthcare or self-driving cars is hindered by their inability to provide safety guarantees or to expose the inner workings of the model in a human understandable form.
We present \Tool, a novel model based on Mixture of Experts, consisting of decision tree experts and a generalized linear model gating function. Thanks to such gating function the model is more expressive than the standard decision tree.
To support non-differentiable decision trees as experts, we formulate a novel training procedure.
In addition, we introduce a hard thresholding version, \MOEHard, in which predictions are made solely by a single expert chosen via the gating function.
Thanks to that property, \MOEHard allows each prediction to be easily decomposed into a set of logical rules in a form which can be easily verified.
While \Tool is a general use model, we illustrate its power in the reinforcement learning setting.
By training \Tool models using an imitation learning procedure on deep RL agents we outperform the previous state-of-the-art technique based on decision trees while preserving the verifiability of the models. Moreover, we show that \Tool can also be used in real-world supervised problems on which it outperforms other verifiable machine learning models. 
\end{abstract}
 
\begin{keyword}
  Verification \sep Deep Learning \sep Reinforcement Learning \sep Mixture of Experts \sep Explainability
\end{keyword}

\end{frontmatter}

\Section{Introduction}
\glsresetall

Deep learning has achieved many recent breakthroughs, in challenging domains such as Go~\cite{SilverETAL16MasteringGO}, and healthcare~\cite{miotto2018deep,esteva2019guide} to name a few.
Encoding state representation via deep neural networks allows \DRL agents to achieve superior performance.
Also it enables development of performant radiology models~\cite{cheng2016computer,cicero2017training,kooi2017large}.
However, the models learned do not provide safety guarantees and are hard to analyze, which hinders their use in safety-critical applications.

An effective recent approach, called \Viper, follows the \Dagger imitation learning procedure~\cite{RossETAL11ImitationLearning} to create a decision tree model mimicking a \DRL agent~\cite{BastaniETAL18VerifiableRL}.
The key advantage of such decision tree models is that they are amenable to verification.
Moreover, they are shown to perform well on environments such as Pong.
However, decision trees are limited to axis perpendicular decision boundaries, which can adversely impact the performance.
In this paper, we alleviate this issue by proposing a model with less restrictions on the geometry of decision boundaries.

We present \Tool (Mixture of Expert Trees), a technique based on \MOE~\cite{JacobsETAL91MOE,JordanXu95MOEConvergence,YukselETAL12TwentyYearsOfMOE}.
\Tool consists of \DT experts and a gating function that determines the weights with which experts are used.
Standard \MOE models can typically use any expert as long as it is a differentiable function of model parameters. %
In this paper we tackle the problem of using non-differentiable decision trees in \MOE context, as a means of obtaining verifiable \DRL agents.
Similar to \MOE training by \EM algorithm, we first observe that \Tool can be trained by interchangeably optimizing the weighted log likelihood for experts (independently from one another) and optimizing the gating function with respect to the obtained experts.
Based on that, we propose a procedure for \DT learning in the specific context of MOE.
To the best of our knowledge we are first to combine standard non-differentiable \DT experts with \MOE approach.

For a gating function, we use a simple generalized linear model with softmax function, which provides a distribution over experts.
While decision boundaries of \DTs are axis-perpendicular, the softmax gating function supports boundaries with hyperplanes of arbitrary orientations, thus improving expressiveness.
We also consider a variant of \Tool model that uses hard thresholding (\MOEHard) which selects just one most likely expert tree.
Since \MOE training algorithm tends to assign a region of space to a single expert ($P(e|r) \approx 1$) anyway, this variant does not suffer in performance, as we empirically demonstrate.
Benefits of \MOEHard compared to the soft version of \Tool are that it
(a) allows for decomposing a decision into a set of logical rules, thus providing means for interpreting the model decisions,
and (b) allows translation to satisfiability modulo theories (SMT)~\footnote{Very roughly, SMT is the problem of determining whether a mathematical formula is satisfiable, and it generalizes the Boolean satisfiability problem (SAT) to more complex formulas.} formulas~\cite{SAThandbook}, thus providing rich opportunities for formal verification using off the shelf SMT solvers~\footnote{SMT solvers are tools designed to solve SMT problems.}, as we demonstrate in the paper.

To employ \Tool in \DRL setting we use the \Dagger imitation learning procedure to mimic \DRL agents.
We evaluate our technique on six different environments: \Cartpole, \Pong, \Acrobot, \Mountaincar, \Lunarlander and \Pendulum.
We show that \Tool achieves better rewards and lower misprediction rates than \Viper.
Finally, we demonstrate how a \Tool policy for \Cartpole can be translated into an SMT formula to verify its properties using the \zt theorem prover~\cite{de2008z3}.
In addition we showed that \Tool can also be used in real-world supervised machine learning problems. We demonstrated that compared to the other verifiable machine learning models (logistic regression, decision trees and support vector classifiers with linear kernels) \Tool achieved much better results.
By improving reliability of AI systems and to a degree improving their interpretability, our work aims at positive societal impact.

In summary, this paper makes the following key contributions:

\begin{enumerate}
\item We propose \Tool, a technique based on \MOE with decision tree experts, and present a learning algorithm to train \Tool models.
\item We create \MOEHard, \Tool version with hard thresholding and softmax gating function which can be translated to an SMT formula amenable for verification and is not hard to interpret in case of small models.
\item We apply \Tool models in the RL setting, evaluate it on different environments and show that they lead to more performant models compared to \Viper decision trees.
\item We apply \Tool models in real-world supervised problems and show that \Tool achieved better results compared to the others verifiable machine learning models.
\end{enumerate}

The remainder of the paper is structured as follows. In section~\ref{sec:Related work} the related work is reviewed. Motivating example to showcase some of the key difference between \Viper and \Tool is presented in section \ref{sec:example}, whereas background methodology is presented in~\ref{sec:Background}. Explanation of \Tool model is given in section~\ref{sec:moet}. Experimental setup and results obtained on different RL environments and supervised datasets are presented in section~\ref{sec:Experiments}. The conclusions are drawn in section~\ref{sec:Conclusion}.
We open source our technique and make it available at: \url{https://github.com/marko-vasic/MoET}.

\Section{Related Work}

\label{sec:Related work}

\textbf{Verifiable Machine Learning:}
RL algorithms are notoriously hard to debug and verify \cite{van2017challenges, amir2021towards}.
A number of techniques has been proposed for enabling verification in RL setting~\cite{zhu2019inductive,kazak2019verifying,verma2019imitation,li2019formal}.
One existing approach synthesizes a program that approximates an RL policy~\cite{zhu2019inductive}.
The program acts as a shield, and their technique coordinates between using the shield program and original policy, which in combination provide safety guarantees.
Instead of using a programmatic policy as a shield, another approach~\cite{verma2019imitation} creates a programmatic policy that can replace neural network policy altogether.
Niu et al. \cite{niu2020toward} provide a general framework that leverages the success of verifiable and safe model-free RL in learning high performance controllers. 
Another system for verification of deep RL agents is presented in \cite{kazak2019verifying}. A hybrid RL agent framework that produces high-level autonomous verifiable behavior for unmanned vehicles is introduced in \cite{wang2019towards}. 
An abstraction approach, based on interval Markov decision processes, that yields probabilistic guarantees on accuracy of
policy's execution, 
and presents techniques to build and solve different kind of control problems 
using abstract interpretation, mixed-integer linear programming, entropy-based refinement, and probabilistic model checking is presented in \cite{bacci2022verified}.

Compared to the other approaches, in this paper we propose a pure machine learning technique that is verifiable and 
applicable even outside of the RL setting.
There has also been recent work on verification of random forests and tree ensembles~\cite{tornblom2020formal,tornblom2018formal}.
Such approaches might be useful in our future work to extend verification from \MOETHard to general \MOET models (which we describe later).

\textbf{Explainable Machine Learning:}
There has been a lot of recent interest in explaining decisions of black-box models ~\cite{guidotti2018survey,doshi2017towards, roscher2020explainable}.
Nowadays, a large set of explainable RL literature is emerging, intended to provide ethical, responsible and trustable algorithms for explaining model outputs of DRL agents \cite{heuillet2021explainability, puiutta2020explainable, wells2021explainable}. Shi et al. \cite{shi2021xpm} proposed XPM -- an explainable RL framework for portfolio management optimization that is based on application of class activation mappings for output explanation. Similarly, Ayala et al. \cite{ayala2021explainable} proposed the introspection-based method for transforming Q-values into probabilities of success, used as the base to explain the agent's decision-making process.
Besides of the explainable RL algorithms, the two most well known algorithms that are commonly used for deep learning models interpretation are  
LIME~\cite{ribeiro2016should} and LORE~\cite{GuidottiETAL18LocalExplanations}.
LIME and LORE explain behavior of a black-box model locally, around an input of interest, by sampling the black-box model around the neighborhood of the input\XComment{using a genetic algorithm (LORE)}, and training a local \DT (or a linear model) over the sampled points.

Another view at \MOET is that it explains behavior of a deep RL agent.
\MOET combines local trees into a global policy by combining local decision trees via a gating function. Inspection of the
trees and the gating might shed light on the agent's decision making. However, we do not focus on this aspect in this paper.

\textbf{Tree-Structured Models:} Tree-Structured models are very attractive type of machine learning algorithms due to low complexity and interpretability \cite{niuniu2010notice, kotsiantis2013decision}. Irsoy et al.~\cite{irsoy2012soft} propose a decision tree model with soft decisions at internal nodes where children are chosen with probabilities given by a sigmoid gating function. However, this reduces the tree's interpretability.
Binary tree-structured hierarchical routing mixture of experts (HRME) model, which has classifiers as non-leaf node experts and simple regression models as leaf node experts, was proposed in~\cite{zhao2019hierarchical}.
Hester and Stone~\cite{hester2013texplore} use random forests in RL setting to build a model of environment from which policy is inferred.

The form of our model can be related to these models, but it is designed with verifiability in mind and we also propose a novel
training procedure suited to that specific model.

\textbf{Knowledge Distillation and Model Compression:} We rely on ideas already explored in fields of model compression~\cite{BuciluaETAL2006ModelCompression} and knowledge distillation~\cite{HintonETAL15Distilling, gou2021knowledge, wang2021knowledge}.
The idea is to use a complex well performing model to facilitate training of a simpler model which might have some other desirable properties (e.g., verifiability and interpretability).
Such practices have been applied to approximate decision tree ensemble by a single tree~\cite{Breiman96BornAgain}. In contrast, we approximate a neural network. Similarly, a neural network can be used to train another neural network~\cite{Furlanello18BornAgainNNs}, but neural networks are hard to interpret and even harder to formally verify.
Such practices have also been applied in the field of reinforcement learning in knowledge and policy distillation \cite{RusuETAL16PolicyDistillation,KoulETAL19FiniteStateRepresentations,ZhangETAL19CausalStateRepresentations, tsantekidis2021diversity, gao2021knowru}, which are similar in spirit to our work, and imitation learning \cite{BastaniETAL18VerifiableRL,RossETAL11ImitationLearning,Abbeel04Apprenticeship,Schaal99Imitation}, which provide a foundation for our work.

\Section{Motivating Example: Gridworld}
\label{sec:example}

\begin{figure*}[!t]
  \centering
  \begin{subfigure}[!t]{0.31\textwidth}
    \centering
    \includegraphics[scale=0.40]{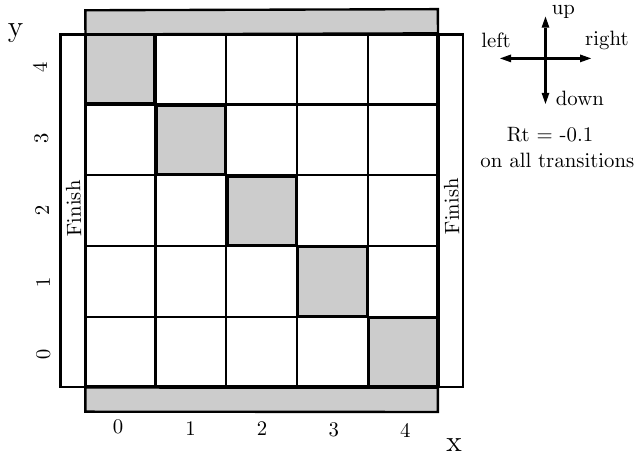}
    \label{fig:gridworld}
  \end{subfigure}
  \begin{subfigure}[!t]{0.30\textwidth}
    \centering
    \includegraphics[scale=0.20]{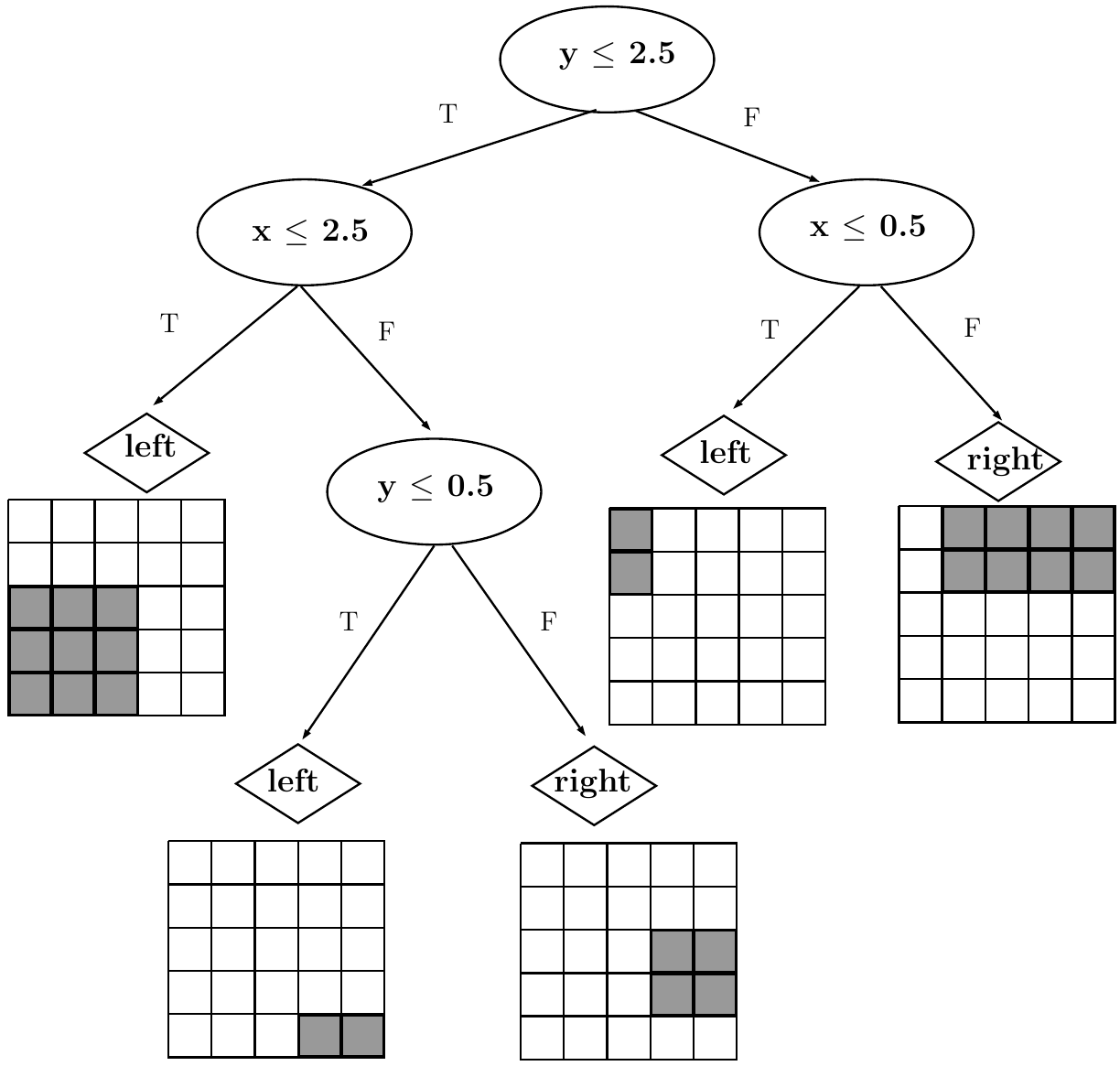}
    \label{fig:gridworld_viper_policy}
  \end{subfigure}
  \begin{subfigure}[!t]{0.33\textwidth}
    \centering
    \includegraphics[scale=0.34]{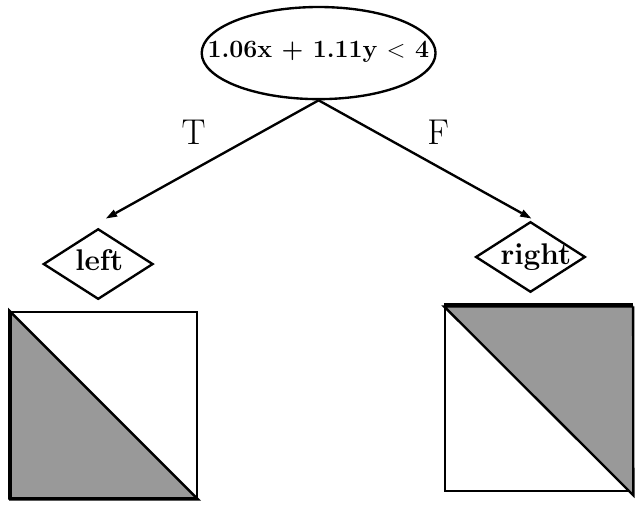}
    \label{fig:gridworld_moet_policy}
  \end{subfigure}
  \caption{
    An illustratory Gridworld environment (left),
    a \Viper policy learned for the environment (middle), and 
    a \Tool policy learned for the environment (right).
  }
\label{fig:motivating-example}
\end{figure*}

We now present a simple motivating example to showcase some of the key
differences between \Viper and \MOET approaches. Consider the $N
\times N$ Gridworld problem shown in \fig~\ref{fig:motivating-example} (for
$N=5$).  The agent is placed at a random position in a grid (except
the walls denoted by filled rectangles) and should find its way out.
To move through the grid the agent can choose to go up, left, right or
down at each time step.  If it hits the wall (gray cell) it stays in the same
position (state).  State is represented using two integer values
($x,y$ coordinates) which range from $(0,0)$---bottom left to
$(N-1,N-1)$---top right.  The grid can be escaped through either left
doors (left of the first column), or right doors (right of the last
column).  A negative reward of $-0.1$ is received for each agent
action (negative reward encourages the agent to find the exit as fast
as possible).  An episode finishes as soon as an exit is reached or if
$100$ steps are made whichever comes first.

The optimal policy ($\pi_*$) for this problem consists of taking the left (right resp.) action for each state below (above resp.) the diagonal.
We used $\pi_*$ as a teacher and imitation learning approach of \Viper to train an interpretable \DT policy that mimics $\pi_*$.
The resulting \DT policy is shown in \fig~\ref{fig:motivating-example}.
The \DT partitions the state space (grid) using lines perpendicular to x and y axes, until it separates all states above diagonal from those below.
This results in a \DT of depth $3$ with $9$ nodes.
On the other hand, the policy learned by \MOET is shown in \fig~\ref{fig:motivating-example}.
The \MOET model with $2$ experts learns to partition the space using the line defined by a linear function $1.06x + 1.11y = 4$ (roughly the diagonal of the grid).
Points on the different sides of the line correspond to two different experts which are themselves \DTs of depth $0$ always choosing to go left (below) or right (above).

\begin{table}
  \caption{
    Size comparison of \MOET and \Viper \DT policies on the Gridworld problem (\fig~\ref{fig:motivating-example}), for different sizes of the square board ($N\times N$).
    The left side of the table presents the depths of obtained models (that perfectly mimic optimal policy) for \MOET and for \Viper (\DTs),
    while the right side presents the number of nodes in these models.
    Both the depth and the number of nodes show that by increasing size of the grid ($N$) size of \MOET models stays constant, while \Viper (\DT) models grow in size.
  }
\small
\label{tbl:gridworld-perf}
\centering
\begin{tabular}{c|cc|cc}
  \toprule
  \TNC{1}{} \vline & \TNC{2}{\HighlightCell{Depth}} & \TNC{2}{\HighlightCell{Nodes}} \\
  \TNC{1}{\HighlightCell{N}} \vline & \TNC{1}{\HighlightCell{\MOETbl}} & \TNC{1}{\HighlightCell{\Viper~\DT}} & \TNC{1}{\HighlightCell{\MOETbl}} & \TNC{1}{\HighlightCell{\Viper~\DT}} \\
  \midrule
  5  & 1 & 3 & 3 & 9 \\
  6  & 1 & 4 & 3 & 11 \\
  7  & 1 & 4 & 3 & 13 \\
  8  & 1 & 4 & 3 & 15 \\
  9  & 1 & 4 & 3 & 17 \\
  10 & 1 & 5 & 3 & 21 \\
  \bottomrule
\end{tabular}
\end{table}

We notice that \DT policy needs much larger depth to represent $\pi_*$ while \MOET can represent it as only one decision step.
Furthermore, with increasing $N$ (size of the grid), complexity of \DT grows, while \MOET complexity stays the same; we empirically confirm this as follows.
For Gridworld sizes $N={5,6,7,8,9,10}$, the depths of obtained \DTs are ${3,4,4,4,4,5}$ and the numbers of their nodes are ${9,11,13,15,17,21}$ respectively.
In contrast, \MOET models of the same complexity and structure as the one shown in \fig~\ref{fig:motivating-example} are learned for all values of $N$. %
We present these results in Table~\ref{tbl:gridworld-perf} for better readability (all policies learned are equivalent to $\pi_*$).

\Section{Background}
\label{sec:Background}
In this section we provide description of two relevant methods we build upon: (1) \Viper, an approach for interpretable imitation learning, and (2) \MOE learning framework.

\textbf{\Viper}.
\Viper algorithm (included in appendix) is an instance of \Dagger imitation learning approach, adapted to prioritize critical states based on Q-values.
Inputs to the \Viper training algorithm are (1) environment $e$ which is an finite horizon ($T$-step) \MDP $(S,A,P,R)$ with states $S$, actions $A$, transition probabilities $P: S \times A \times S \to[0, 1]$, and rewards $R: S \to \mathbb{R}$; (2) teacher policy $\pi_t: S \to A$; (3) its Q-function $Q^{\pi_t}: S \times A \to \mathbb{R}$ and (4) number of training iterations $N$.
Distribution of states after $T$ steps in environment $e$ using a policy $\pi$ is $d^{(\pi)}(e)$ (assuming randomly chosen initial state).
\Viper uses the teacher as an oracle to label the data (states with actions).
It initially uses teacher policy to sample trajectories (states) to train a student (\DT) policy. It then uses the student policy to generate more trajectories.
\Viper samples training points from the collected dataset $D$ giving priority to states $s$ having higher importance $I(s)$,
where $ I(s) = \max_{a \in A}Q^{\pi_t}(s,a) - \min_{a \in A}Q^{\pi_t}(s,a) $. %
This sampling of states leads to faster learning and shallower \DTs.
The process of sampling trajectories and training students is repeated for number of iterations $N$, and the best student policy is chosen using reward as the criterion.

\XComment{As already mentioned, \Viper is an instance of \Dagger algorithm,
with a modification of using importance sampling based on action importance measured by \Qvalue.
We use the same imitation learning approach in \Tool, collecting the training points in a same manner,
with a difference that we do not train a \DT for a student policy but we train a \Tool model.}

\textbf{Mixture of Experts}.
\MOE is an ensemble model~\cite{JacobsETAL91MOE,JordanXu95MOEConvergence,YukselETAL12TwentyYearsOfMOE} that consists of expert networks and a gating function.
Gating function divides the input (feature) space into regions for which different experts are specialized and responsible.
\MOE is flexible with respect to the choice of expert models as long as they are differentiable functions of model parameters (which is not the case for \DTs).

In \MOE framework, probability of outputting $\vec{y} \in {\rm I\!R}^m$ given an input $\vec{x} \in {\rm I\!R}^n$ is given by:
\BeginEquation
P(\vec{y} | \vec{x}, \vec{\theta}) = \sum_{i=1}^{E} P(i | \vec{x}, \vec{\theta}_g) P(\vec{y} | \vec{x}, \vec{\theta}_i) = \sum_{i=1}^{E} g_i(\vec{x}, \vec{\theta}_g) P(\vec{y} | \vec{x}, \vec{\theta}_i)
  \label{eq:MoE}
\end{equation}
where $E$ is the number of experts, $g_i(\vec{x}, \vec{\theta}_g)$ is the probability of choosing the expert $i$ (given input $\vec{x}$), $P(\vec{y} | \vec{x}, \vec{\theta}_i)$ is the probability of expert $i$ producing output $\vec{y}$ (given input $\vec{x}$).
Learnable parameters are $\vec{\theta}=(\vec{\theta}_g,\vec{\theta}_e)$, where $\vec{\theta}_g$ are parameters of the gating function and $\vec{\theta}_e = (\vec{\theta}_1, \vec{\theta}_2, ..., \vec{\theta}_E)$ are parameters of the experts. Gating function can be modeled using a softmax function over a set of linear models. Let $\vec{\theta}_g$ consist of parameter vectors $(\vec{\theta}_{g1},\ldots,\vec{\theta}_{gE})$, then the gating function can be defined as
$
  g_i(\vec{x}, \vec{\theta}_g) = \nicefrac{\exp(\vec{\theta}^T_{gi}\vec{x})}{\sum_{j=1}^{E} \exp(\vec{\theta}^T_{gj}\vec{x})}
$
.

\XComment{In a case of regression, normal distribution is typically assumed for experts. Thus, $\vec{\theta_i} = \{\vec{\mu}_i, \vec{\Sigma}_i\}$; where $\vec{\mu_i}$ is the expectation, and $\vec{\Sigma}_i$ is a covariance matrix.
Therefore, probability of expert $i$ outputting $\vec{y}$ is given by: $P(\vec{y} | \vec{x}, \vec{\theta}_i) = {\cal N}(\vec{y} | \vec{\mu}_i, \vec{\Sigma}_i)$. The output of \MOE is gate-weighted average of expert outputs.}

In the case of classification, an expert $i$ outputs a vector $\vec{y}_i$ of length $C$, where $C$ is the number of classes.
Expert $i$ associates a probability to each output class $c$ (given by $\vec{y}_{ic}$) using the gating function.
Final probability of a class $c$ is a gate weighted sum of $\vec{y}_{ic}$ for all experts $i \in {1, 2, ..., E}$.
This creates a probability vector $\vec{y} = (y_1, y_2, ..., y_C)$, and the output of \MOE is $\argmax_{i} \vec{y}_i$. %

\MOE is commonly trained using an \EM algorithm, where instead of direct optimization of the likelihood one performs optimization of an auxiliary function $\hat{L}$ defined in a following way. Let $z$ denote the expert chosen for instance $\vec{x}$. Then joint likelihood of $\vec{x}$ and $z$ can be considered. Since $z$ is not observed in the data, log likelihood of samples $(\vec{x},z,\vec{y})$ cannot be computed, but instead expected log likelihood can be considered, where expectation is taken over $z$.
Since the expectation has to rely on some distribution of $z$, in the iterative process, the distribution with respect to the current estimate of parameters $\theta$ is used.
More precisely function $\hat{L}$ is defined by \cite{JordanXu95MOEConvergence}:
\BeginEquation
  \hat{L}(\vec{\theta},\vec{\theta}^{(k)}) = \mathbb{E}_z[\log P(\vec{x},z,\vec{y})|\vec{x},\vec{y},\vec{\theta}^{(k)}] = \int P(z|\vec{x},\vec{y},\vec{\theta}^{(k)})\log P(\vec{x},z,\vec{y})dz
\end{equation}
where $\vec{\theta}^{(k)}$ is the estimate of parameters $\vec{\theta}$ in iteration $k$. Then, for a specific sample $D=\{(\vec{x}_i,\vec{y}_i)\ |\ i=1,\ldots,N\}$, the following formula can be derived \cite{JordanXu95MOEConvergence}:
\BeginEquation
  \hat{L}(\vec{\theta},\vec{\theta}^{(k)}) = \sum_{i=1}^N\sum_{j=1}^Eh^{(k)}_{ij}\log g_j(\vec{x}_i,\vec{\theta}_g) + \sum_{i=1}^N\sum_{j=1}^Eh^{(k)}_{ij}\log P(\vec{y}_i|\vec{x}_i,\vec{\theta}_j)
  \label{eq:q1}
\end{equation}
where it holds
\BeginEquation
h^{(k)}_{ij}=\frac{g_j(\vec{x}_i,\vec{\theta}^{(k)}_g)P(\vec{y}_i|\vec{x}_i,\vec{\theta}^{(k)}_j)}{\sum_{l=1}^Eg_l(\vec{x}_i,\vec{\theta}^{(k)}_g)P(\vec{y}_i|\vec{x}_i,\vec{\theta}^{(k)}_l)}
\label{eq:q2}
\end{equation}

\Section{Mixture of Expert Trees}
\label{sec:moet}

In this section we explain the adaptation of original \MOE model to mixture of decision trees, and present both training and inference algorithms.

Considering that coefficients $h^{(k)}_{ij}$ (\Equat~\ref{eq:q2}) are fixed with respect to $\vec{\theta}$ and that in \Equat~\ref{eq:q1} the gating part (first double sum) and each expert part depend on disjoint subsets of parameters $\vec{\theta}$, training can be carried out by interchangeably optimizing the weighted log likelihood for experts (independently from one another) and optimizing the gating function with respect to the obtained experts.
The training procedure for \Tool, described by Algorithm~\ref{alg:training}, is based on this observation.
First, the parameters of the gating function are randomly initialized (line~\ref{alg:training:line:gating-init}).
Then the experts are trained one by one. Each expert $j$ is trained on a dataset $D_w$ of instances weighted by coefficients $h^{(k)}_{ij}$ (line~\ref{alg:training:line:weighting}), by applying specific \DT learning algorithm (line~\ref{alg:training:line:dtlearning}) that we adapted for \MOE context (described below).
After the experts are trained, an optimization step is performed (line~\ref{alg:training:line:gating-optim}) in order to increase the gating part of \Equat~\ref{eq:q1}.
At the end, the parameters are returned (line~\ref{alg:training:line:return}).

Our tree learning procedure is as follows. Our technique modifies original \MOE algorithm in that it uses \DTs as experts.
The fundamental difference with respect to traditional model comes from the fact that \DTs do not rely on explicit and differentiable loss function which can be trained by gradient descent or Newton's methods.
Instead, due to their discrete structure, they rely on a specific greedy training procedure.
Therefore, the training of \DTs has to be modified in order to take into account the attribution of instances to the experts given by coefficients $h^{(k)}_{ij}$, sometimes called {\em responsibility} of expert $j$ for instance $i$.
If these responsibilities were hard, meaning that each instance is assigned to strictly one expert, they would result in partitioning the feature space into disjoint regions belonging to different experts.
On the other hand, soft responsibilities are fractionally distributing each instance to different experts.
The higher the responsibility of an expert $j$ for an instance $i$, the higher the influence of that instance on that expert's training.
In order to formulate this principle, we consider which way the instance influences construction of a tree. First, it affects the impurity measure computed when splitting the nodes and second, it influences probability estimates in the leaves of the tree. We address these two issues next.

A commonly used impurity measure to determine splits in the tree is the Gini index.
Let $U$ be a set of indices of instances assigned to the node for which the split is being computed and $D_U$ set of corresponding instances.
Let categorical outcomes of $y$  be $1,\ldots,C$, and for $l=1,\ldots,C$ let denote $p_l$ as a fraction of instances in $D_U$ for which it holds $y=l$.
More formally
{$p_l=\frac{\sum_{i\in U}I[y_i=l]}{|U|}$},
where $I$ denotes indicator function of its argument expression and equals $1$ if the expression is true.
Then the Gini index $G$ of the set $D_U$ is defined by:
$G(p_1,\ldots,p_C)=1-\sum_{l=1}^Cp^2_l $.
Considering that the assignment of instances to experts are fractional as defined by responsibility coefficients $h^{(k)}_{ij}$ (which are provided to tree fitting function as weights of instances computed in line~\ref{alg:training:line:weighting} of the algorithm), this definition has to be modified in that the instances assigned to the node should not be counted, but instead, their weights should be summed. Hence, we propose the following definition:
\BeginEquation
\small
\hat{p}_l=\frac{\sum_{i\in U}I[y_i=l]h^{(k)}_{ij}}{\sum_{i\in U}h^{(k)}_{ij}}
\label{eq:probabilities}
\end{equation}
and compute the Gini index for the set $D_U$ as $G(\hat{p}_1,\ldots,\hat{p}_C)$. 
Similar modification can be performed for other impurity measures (such as entropy) relying on distribution of outcomes of a categorical variable.
Note that while the instance assignments to experts are soft, instance assignments to nodes within an expert are hard, meaning sets of instances assigned to different nodes are disjoint.
Probability estimate for $\vec{y}$ in the leaf node is usually performed by computing fractions of instances belonging to each class.
Instead of such estimates, again, we use estimates $\hat{p}_l$ defined by \Equat~\ref{eq:probabilities}. 
Hence, the estimates of probabilities $P(\vec{y}|\vec{x},\theta^{(k)}_j)$ needed by \MOE are defined.
In Algorithm~\ref{alg:training}, function $fit\_tree$ performs decision tree training using the above modifications.

\setlength{\textfloatsep}{1ex}
\begin{algorithm}[!t]
\caption{\Tool training.}\label{alg:training}
\begin{algorithmic}[1]
\Procedure{\Tool(Data $\{(\vec{x}_i,\vec{y}_i)\ |\ i=1,\ldots,N\}$, Epochs $N_E$, Number of Experts $E$)}{}
\State $\vec{\theta_g} \gets initialize()$ \label{alg:training:line:gating-init}
\For {$k \gets 1$ to $N_E$}
\For {$j \gets 1$ to $E$}
\State $D_w \gets \left\{\left(\vec{x}_i,\vec{y}_i,\frac{g_j(\vec{x}_i,\vec{\theta}_g)P(\vec{y}_i|\vec{x}_i,\vec{\theta}_j)}{\sum_{e=1}^Eg_e(\vec{x}_i,\vec{\theta}_g)P(\vec{y}_i|\vec{x}_i,\vec{\theta}_e)}\right)\ |\ i=1,\ldots,N\right\}$ \label{alg:training:line:weighting}
\State $\vec{\theta}_j \gets fit\_tree(D_w)$ \label{alg:training:line:dtlearning}
\EndFor
\State $\vec{\theta}_g \gets \vec{\theta}_g+\lambda \nabla_{\theta'}\sum_{i=1}^N\sum_{j=1}^E\left[\frac{g_j(\vec{x}_i,\vec{\theta}_g)P(\vec{y}_i|\vec{x}_i,\vec{\theta}_j)}{\sum_{e=1}^Eg_e(\vec{x}_i,\vec{\theta}_g)P(\vec{y}_i|\vec{x}_i,\vec{\theta}_e)}\log g_j(\vec{x}_i,\vec{\theta}')\right]$ \label{alg:training:line:gating-optim}
\EndFor
\State \Return $\vec{\theta}_g, (\theta_1,\ldots,\theta_E)$ \label{alg:training:line:return}
\EndProcedure
\end{algorithmic}
\end{algorithm}

We consider two ways to perform inference with respect to the obtained model. First one which we call \Tool, is performed by maximizing $P(\vec{y}|\vec{x},\vec{\theta})$ with respect to $\vec{y}$ where this probability is defined by \Equat~\ref{eq:MoE}. The second way, which we call \MOEHard, performs inference as
$ \argmax_\vec{y}P(\vec{y}|\vec{x},\vec{\theta}_{\argmax_j g_j(x,\vec{\theta}_g)}) $,
meaning that we only rely on the most probable expert.

\textbf{Adaptation of \Tool to imitation learning}.
We integrate \Tool model into imitation learning approach of \Viper by substituting training of \DT with the \Tool training procedure.

\textbf{Verifiability by translating \MOET to SMT}.
We define a translation of \MOETHard models to SMT formulas, which opens a range of possibilities for validating and interpreting the model using automated reasoning tools.
SMT formulas provide a rich means of logical reasoning, where a user can query the solver with questions such as:
``What inputs do the two models differ on?'', or ``What is the closest input to the given input using which model makes a different prediction?'', or ``Are the two models equivalent?'', or ``Are the two models equivalent in respect to the output class C?''.
Answers to such questions can help better understand and compare models in a rigorous way.
Also note that the symbolic reasoning of the gating function and decision trees allows construction of SMT formulas that are readily handled by off-the-shelf tools, whereas direct SMT encoding of neural networks do not scale for any reasonably sized network because of the need for non-linear arithmetic reasoning.

We show the translation of \MOET policy to SMT constraints for verifying policy properties. We present an example translation of \MOET policy on \Cartpole environment with the same property specification that was proposed for verifying \Viper policies~\cite{BastaniETAL18VerifiableRL}.
The goal in \Cartpole is to keep the pole upright, which can be encoded as a formula:
\[ \small \psi\equiv s_0\in S_0\wedge\bigwedge_{t=1}^{\infty}|\phi(f(s_{t-1},\pi(s_{t-1})))|\le y_0 \]
where $s_i$ represents state after $i$ steps, $\phi$ is the deviation of pole from the upright position.
In order to encode this formula it is necessary to encode the transition function $f(s, a)$ which models environment dynamics:
given a state and action it returns the next state of the environment.
Also, it is necessary to encode the policy function $\pi(s)$ that for a given state returns action to perform. %
There are two issues with verifying $\psi$: (1) infinite time horizon; and (2) the nonlinear transition function $f$.
To solve this problem, Bastani et al.~\cite{BastaniETAL18VerifiableRL} use a finite time horizon $T_{max}=10$ and linear approximation of the dynamics.
We make the same assumptions.

To encode $\pi(s)$ we need to translate both the gating function and \DT experts to logical formulas. Since the gating function in \MOETHard uses exponential function, it is difficult to encode the function directly in \zt as SMT solvers do not have efficient decision procedures to solve non-linear arithmetic. The direct encoding of exponentiation therefore leads to prohibitively complex \zt formulas. We exploit the following simplification of the gating function that is sound when hard prediction is used:
\BeginEquation
\small
  e = \argmax_i\left(\frac{\exp(\vec{\theta}^T_{gi}\vec{x})}{\sum_{j=1}^{E} \exp(\vec{\theta}^T_{gj}\vec{x})}\right) \\ 
  = \argmax_i(\exp(\vec{\theta}^T_{gi}\vec{x})) \\
  = \argmax_i(\vec{\theta}^T_{gi}\vec{x})
\end{equation}
First simplification is possible since the denominators of the gating functions are same for all experts, and second is due to the monotonicity of the exponential function. 
We use the same \DT encoding as in \Viper. To verify that $\psi$ holds we need to show that $\lnot \psi$ is unsatisfiable.
In the experimental evaluation we run the verification with our \MOETHard policies and show that $\lnot \psi$ is indeed unsatisfiable.

\textbf{Expressiveness}.
\DTs make their decisions by partitioning the feature space into regions which have borders perpendicular to coordinate axes.
To approximate borders that are not perpendicular to coordinate axes very deep trees are often necessary.
\MOEHard mitigates this shortcoming by exploiting hard softmax partitioning of the feature space using borders which are still hyperplanes, but need not be perpendicular to coordinate axes (see Section~\ref{sec:example}), which improves the expressiveness.

\textbf{Interpretability}.
While we do not focus on interpretability in this work, it is useful to note that \MOEHard models do exhibit some interpretability properties.
A \MOEHard model is a combination of a linear model and several decision tree models. Only a single \DT is used for each prediction (instead of weighted average),
which facilitates interpretability. If the models are small (e.g, depth $\leq$ $10$) and include small number of features, a person can easily simulate and understand the model.
These observations resonate with several points about interpretability made in~\cite{Lipton2016Mythos}

\textbf{Limitations}.
Our work tries to strike a balance between expressiveness, which allows for more performant models, and verifiability, which allows for more reliable models.
Therefore, while being more expressive than decision trees, \Tool still has limited expressiveness compared to deep learning models, which is a price paid for easier verifiability.

\setlength{\textfloatsep}{\textfloatsepsave}

\Section{Evaluation}
\label{sec:Experiments}
We first discuss \DRL agents we use as a starting point in the imitation learning.
Second, we explore the performance capabilities of \Viper by finding decision tree depths at which the performance saturates---cannot be improved by increasing the depth further.
Then, after ensuring that we explored the useful space of configurations for \Viper, we pick the best performing \Viper models and compare them with the best performing \MOET models to quantitatively compare the two.
Finally, we re-evaluate performance of the models to evaluate how well they generalize.
Also, we verify \MOETH policies on \Cartpole environment and visually compare the expressiveness of different policies. Eventually, we presented that \Tool can be also successfully applied in real-world supervised learning problems.

\textbf{\DRL agents}.
We use following \OpenAIGym environments in our evaluation: \Cartpole, \Acrobot, \Mountaincar, \Lunarlander, \Pong and \Pendulum (description of the environments is included in the appendix).
For \DRL agents, we use a policy gradient model in \Cartpole, a \DQN~\cite{MnihETAL15DRL} in \Pong, and dueling \DQN~\cite{wang2015dueling} in the other environments (training hyperparameters provided in the appendix).
We train \MOET and \Viper policies by mimicking the agents.
The rewards (total return during an episode) obtained by the \DRL agents on \Cartpole, \Acrobot, \Mountaincar, \Lunarlander, \Pong and \Pendulum are $200.00$, $-68.60$, $-105.27$, $190.90$, $21.00$ and $-158.13$, respectively.
Rewards are averaged across $100$ ($250$ in \Cartpole) runs (episodes).

\begin{figure*}[!t]
\begin{subfigure}[!t]{0.5\textwidth}
  \centering
  \includegraphics[scale=0.4]{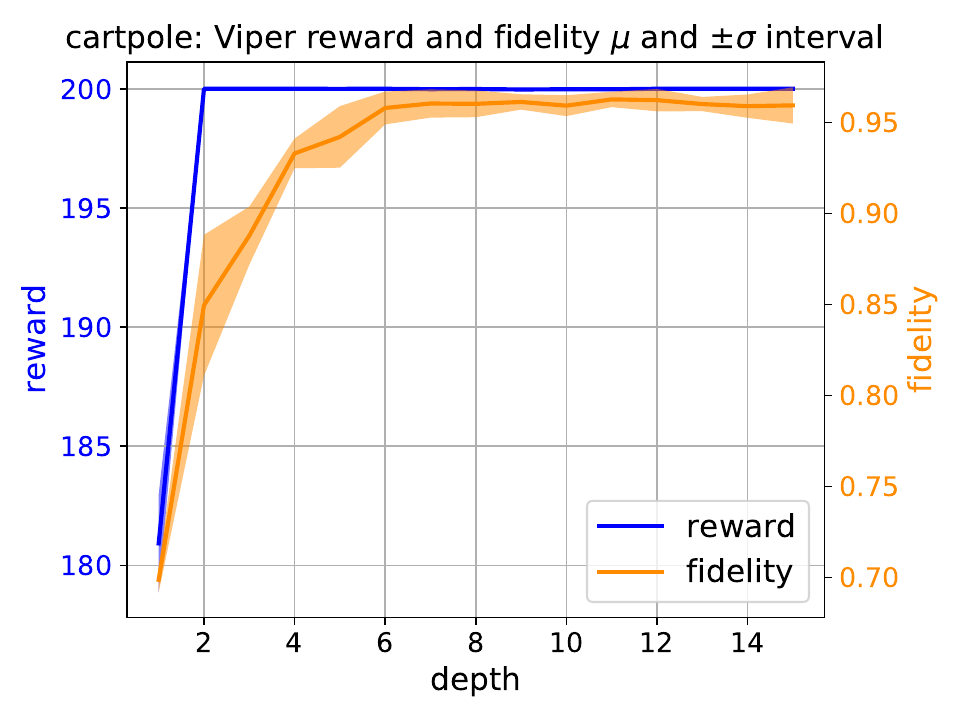}
\end{subfigure}
\begin{subfigure}[!t]{0.5\textwidth}
  \centering
  \includegraphics[scale=0.4]{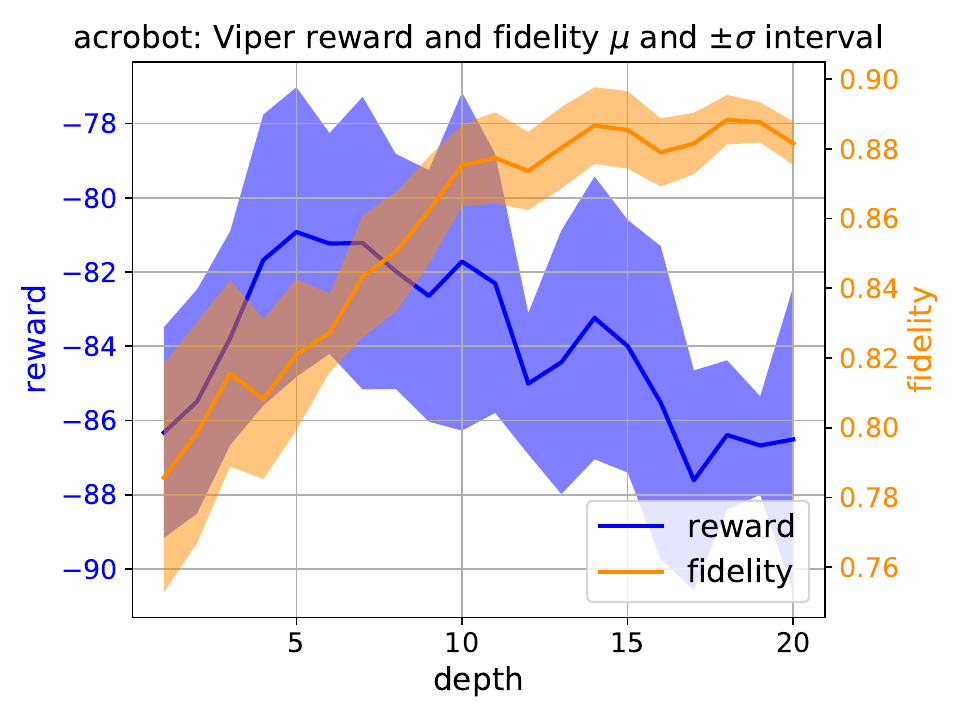}
\end{subfigure}
\begin{subfigure}[!t]{0.5\textwidth}
  \centering
  \includegraphics[scale=0.4]{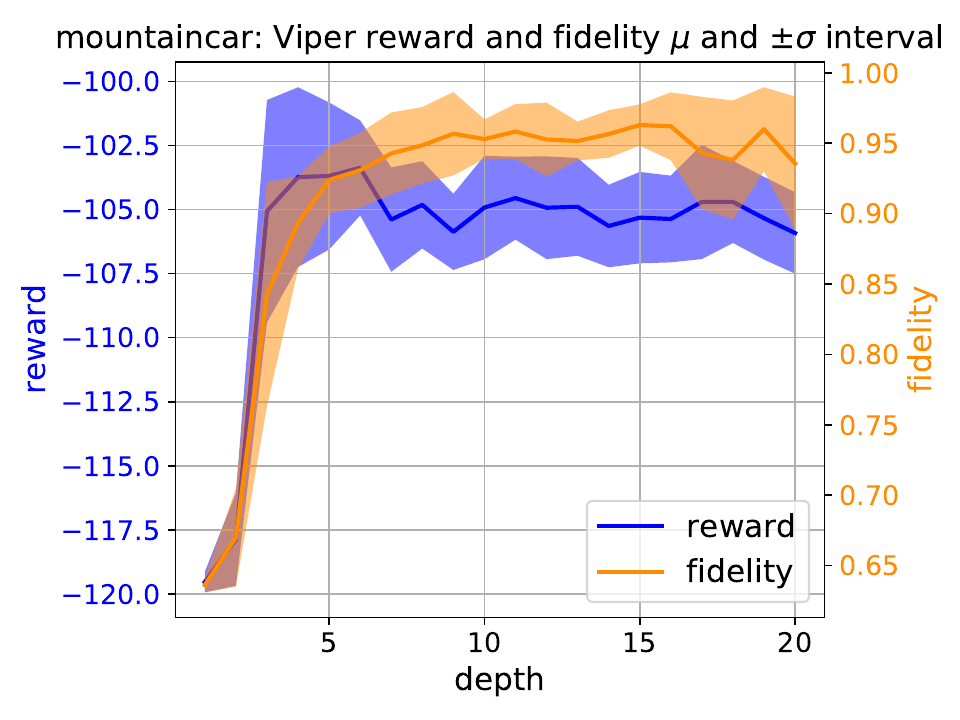}
\end{subfigure}
\begin{subfigure}[!t]{0.5\textwidth}
  \centering
  \includegraphics[scale=0.4]{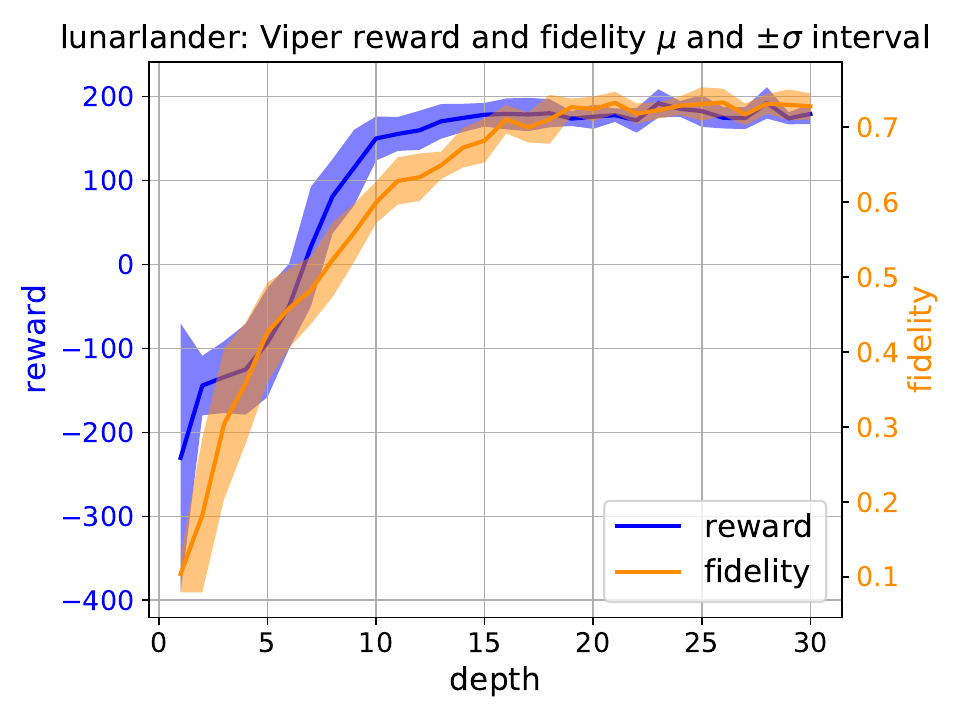}
\end{subfigure}
\begin{subfigure}[!t]{0.5\textwidth}
  \centering
  \includegraphics[scale=0.4]{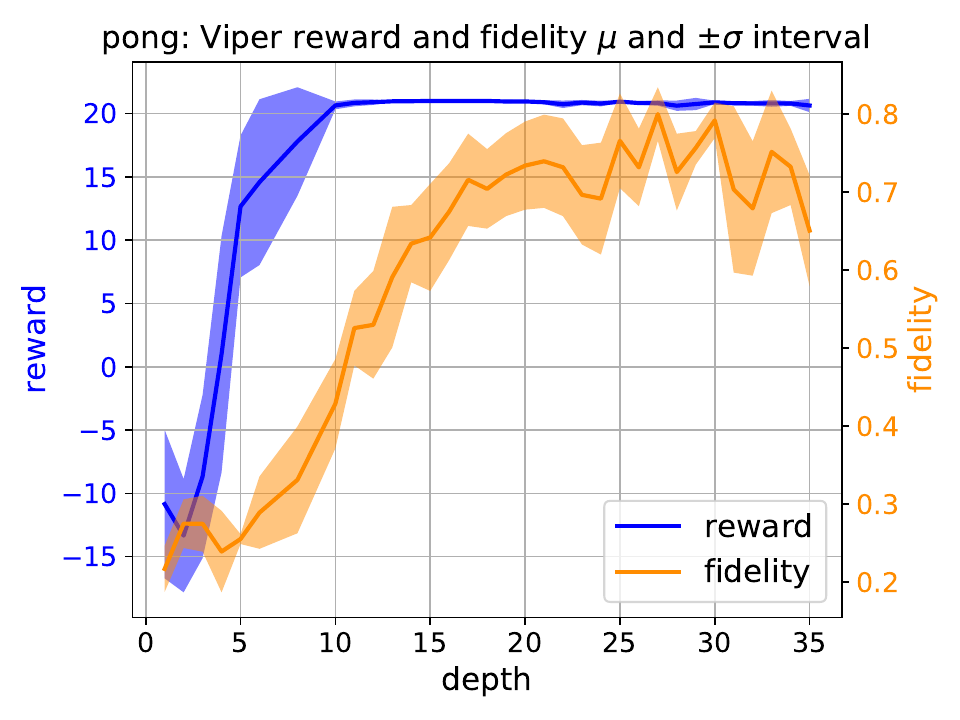}
\end{subfigure}
\begin{subfigure}[!t]{0.5\textwidth}
  \centering
  \includegraphics[scale=0.4]{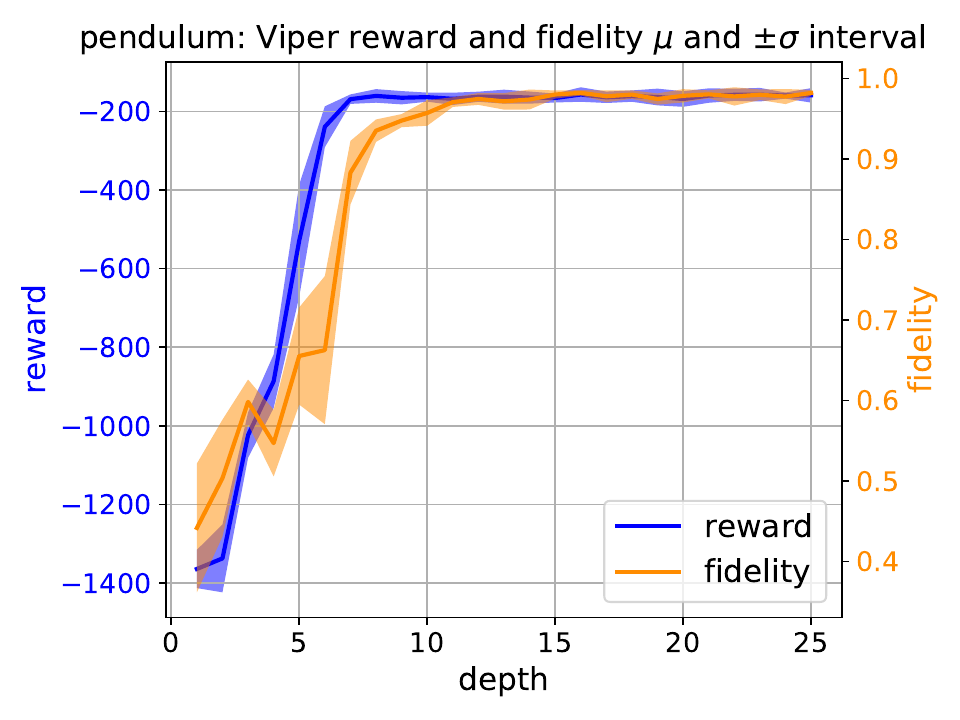}
\end{subfigure}
\caption{
  \textbf{Performance saturation of \Viper}.
  Multiple models are trained for a single maximum depth of \Viper decision trees,
  while maximum depth is incrementally increased,
  showing the mean value and standard deviation of reward and fidelity with respect to the depth.
  These results inform when \Viper performance saturates, i.e., reaches a point upon which increasing maximum depth is not helpful anymore, we call that point performance saturation depth.
}
\label{fig:viper_performance}
\end{figure*}

\textbf{Performance saturation of \Viper}.
We first examine performance capabilities of \Viper, i.e., answer the question of when the performance saturates, by examining performance of decision trees of gradually increased maximum depth (Figure~\ref{fig:viper_performance}).
For each depth we train multiple \Viper models and show performance trends in terms of reward and fidelity.
By reward we mean cumulative reward achieved during an episode,
while fidelity represents percent of times a student performs the same action as its teacher (\DRL agent).
Achieving high reward indicates that a student is performing well, while high fidelity indicates that the student policy is close to the teacher's.
We ensure to train at least $5$ different \Viper models for each depth.\footnote{
  We train at least $5$ \Viper models for each subject and maximum depth value.
  Due to the computational limitations actual number of \Viper models trained varies across environments:
  \Cartpole $\in [35,70]$, \Acrobot $\in [10,70]$, \Mountaincar $\in [10,70]$, \Lunarlander $\in [10,70]$, \Pong $\in [5,24]$ and \Pendulum $\in \{10\}$.
}
Using the performance trend plots we infer when \Viper performance saturates, i.e., reaches a depth after which further increasing maximum depth does not help.
Performance saturation depths for \Cartpole, \Acrobot, \Mountaincar, \Lunarlander, \Pong and \Pendulum are \CartpoleSaturationDepth, \AcrobotSaturationDepth, \MountaincarSaturationDepth, \LunarlanderSaturationDepth, \PongSaturationDepth and \PendulumSaturationDepth, respectively.
Identifying the performance saturation points for \Viper is helpful in identifying the overall best performing \Viper model,
thus giving confidence during comparison with \MOET models that we explored the useful space of \Viper configurations.

\begin{figure*}[!t]
\begin{subfigure}[!t]{0.5\textwidth}
  \centering
  \includegraphics[scale=0.4]{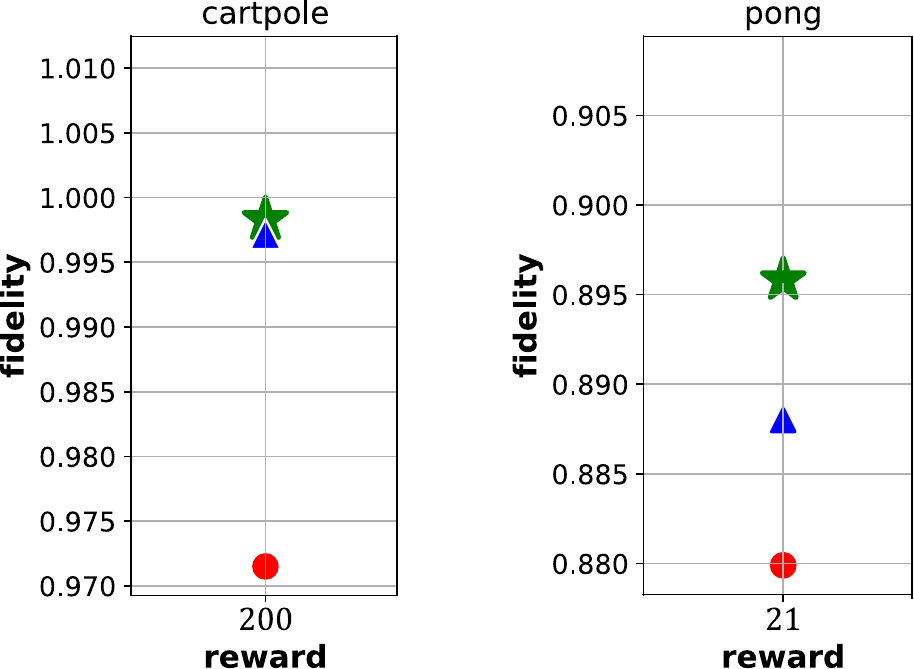}
\end{subfigure}
\begin{subfigure}[!t]{0.5\textwidth}
  \centering
  \includegraphics[scale=0.4]{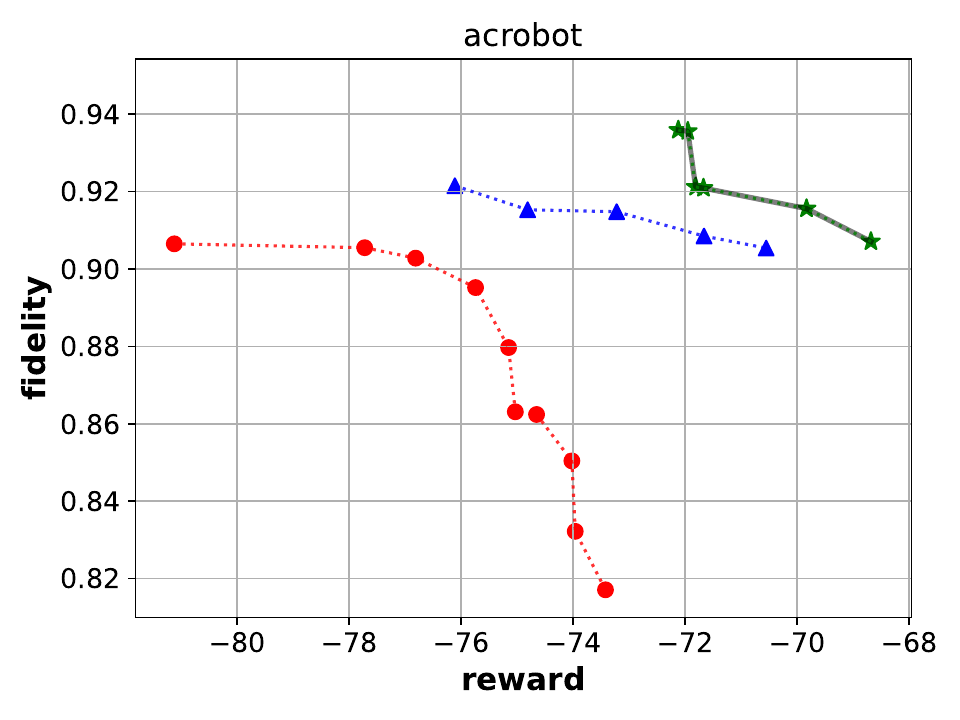}
\end{subfigure}
\begin{subfigure}[!t]{0.5\textwidth}
  \centering
  \includegraphics[scale=0.4]{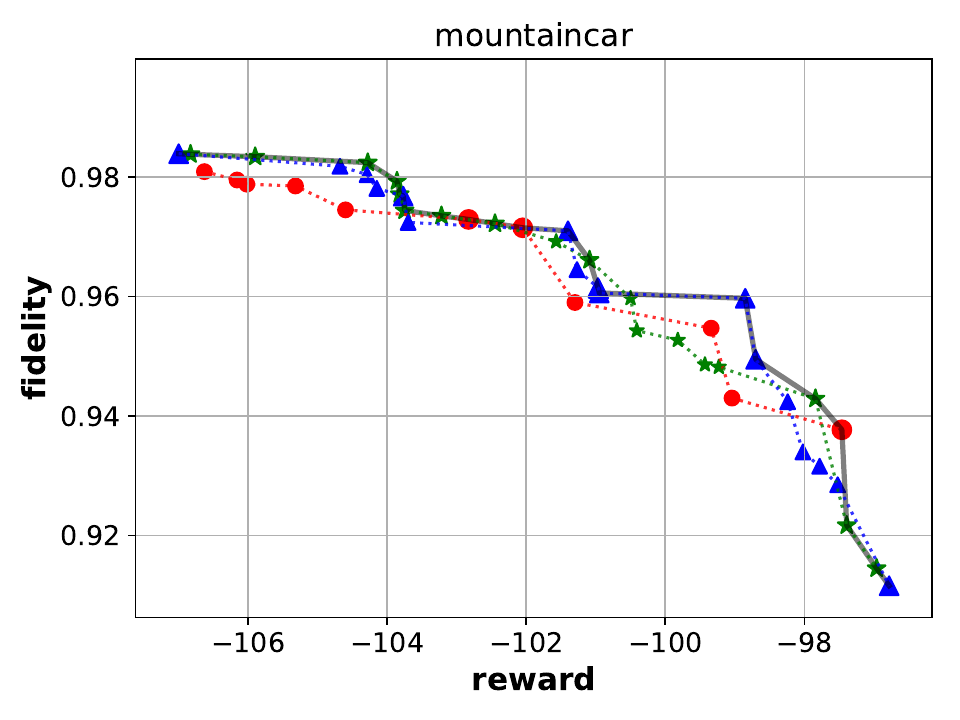}
\end{subfigure}
\begin{subfigure}[!t]{0.5\textwidth}
  \centering
  \includegraphics[scale=0.4]{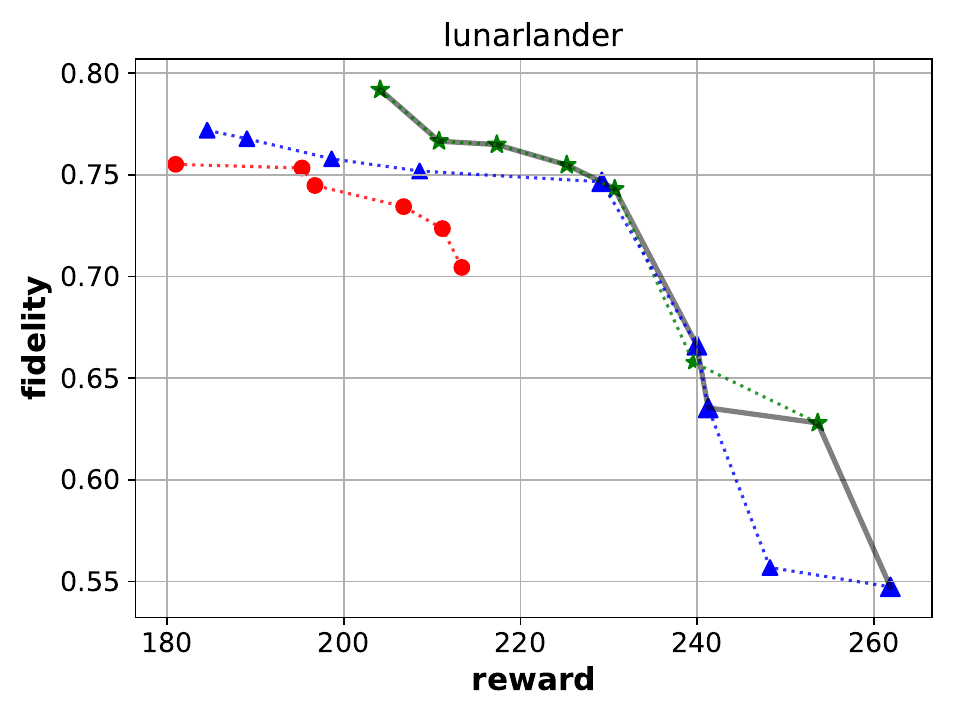}
\end{subfigure}
\begin{subfigure}[!t]{0.5\textwidth}
  \centering
  \includegraphics[scale=0.4]{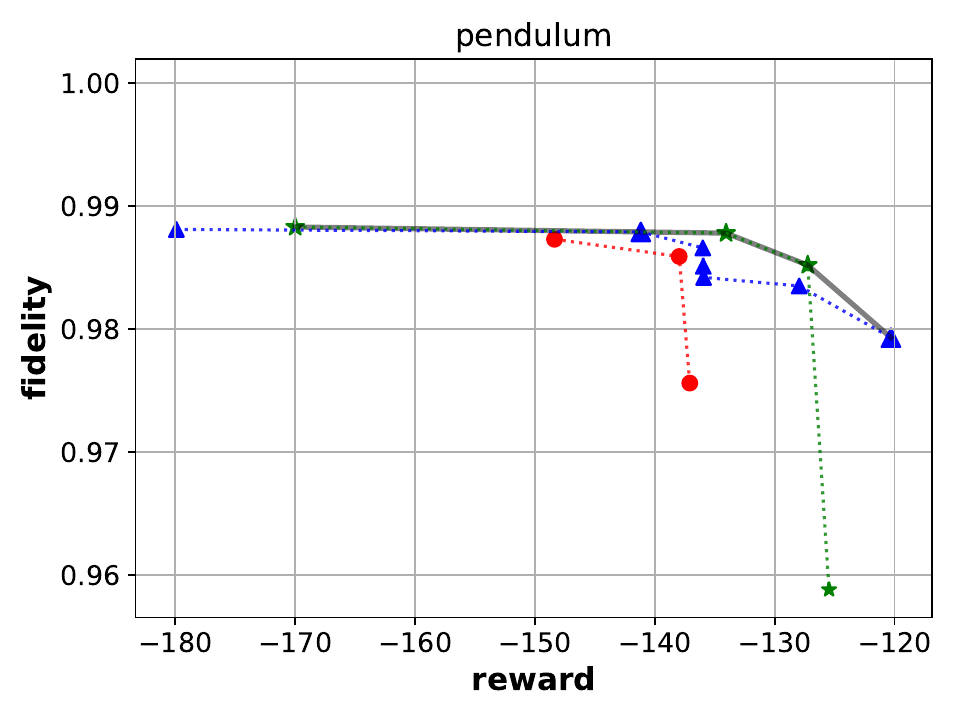}
\end{subfigure}
\begin{subfigure}[!t]{0.5\textwidth}
  \centering
  \includegraphics[scale=0.45]{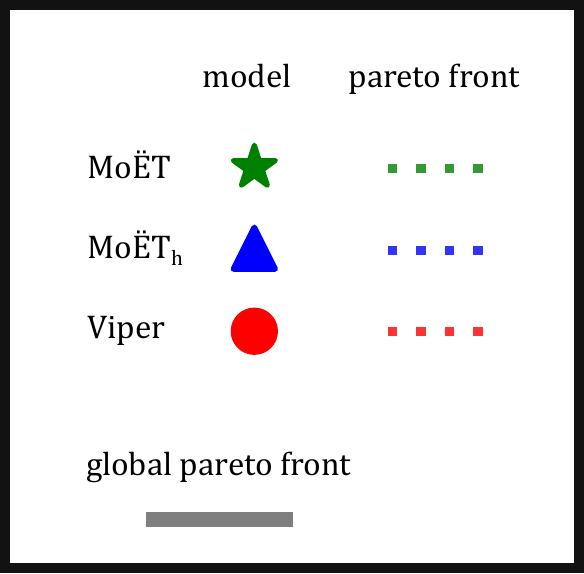}
\end{subfigure}
\caption{
  \textbf{Best performing \Viper, \MOET and \MOETH models}.
  Pareto fronts (in respect to the reward and fidelity) are identified separately for \Viper, \MOET and \MOETH models.
  Global Pareto fronts are shown with points connected by a gray solid line.
}
\label{fig:pre-evaluation}
\end{figure*}

\textbf{Best performing \Viper, \MOET and \MOETH models}.
We next compare \Viper, \MOET and \MOETH models by visualizing their Pareto fronts with respect to the reward and fidelity (Figure~\ref{fig:pre-evaluation}).
Pareto front of a set of models consists of all models from that set which are not dominated by any other model from the set in terms of reward or fidelity.
In other words, every model dominated by another model in terms of both metrics is not considered.
From the set of all \Viper models trained for different maximum depths (from depth $1$ to the saturation depth) we select models on the Pareto front.
Similar is done for \MOET and \MOETH which we trained for different number of experts and expert depths (information about configurations used is provided in the appendix).
A global Pareto front (best models across all architectures) is shown with points connected by a black solid line.

By inspecting the results we notice that in the case of \Cartpole, all $3$ models achieve maximum reward ($200$), however fidelity is significantly higher in the case of \MOET and \MOETH (over $99\%$ compared to $97\%$).
Also, it is interesting to note that both \MOET and \MOETH models on the Pareto front consist of $2$ experts of depth $0$, while the \Viper model on the Pareto front is a decision tree of depth $6$.
In the case of \Acrobot, we notice that \MOET models dominate \MOETH and \Viper models, and that \MOETH models dominate \Viper models.
Thus, both \MOET and \MOETH models achieve higher reward and fidelity over \Viper models.
In the case of \Mountaincar, the global Pareto front contains some \Viper models, but mostly \MOET and \MOETH dominate.
Furthermore, models exhibiting the highest reward as well as fidelity are \MOET and \MOETH models.
In the case of \Lunarlander, both \MOET and \MOETH dominate \Viper models.
A \MOETH model achieves the maximum reward of over $260$ while a \Viper model achieves the maximum reward of around $215$.
Furthermore, both \MOET and \MOETH models achieve better fidelity compared to \Viper.
In the case of \Pong, all $3$ models achieve maximum reward ($21$), however fidelity is higher for \MOET and \MOETH.
In the case of \Pendulum, \MOET and \MOETH models achieve better maximum reward, while maximum fidelity is about equal for all the models.
Note that for a given fidelity score, \MOET and \MOETH are advantageous to \Viper.
Scores of the points on the global Pareto front are presented in a tabular form in \ref{sec:eval-appendix}.

\begin{figure*}[!t]
\begin{subfigure}[!t]{0.5\textwidth}
  \centering
  \includegraphics[scale=0.4]{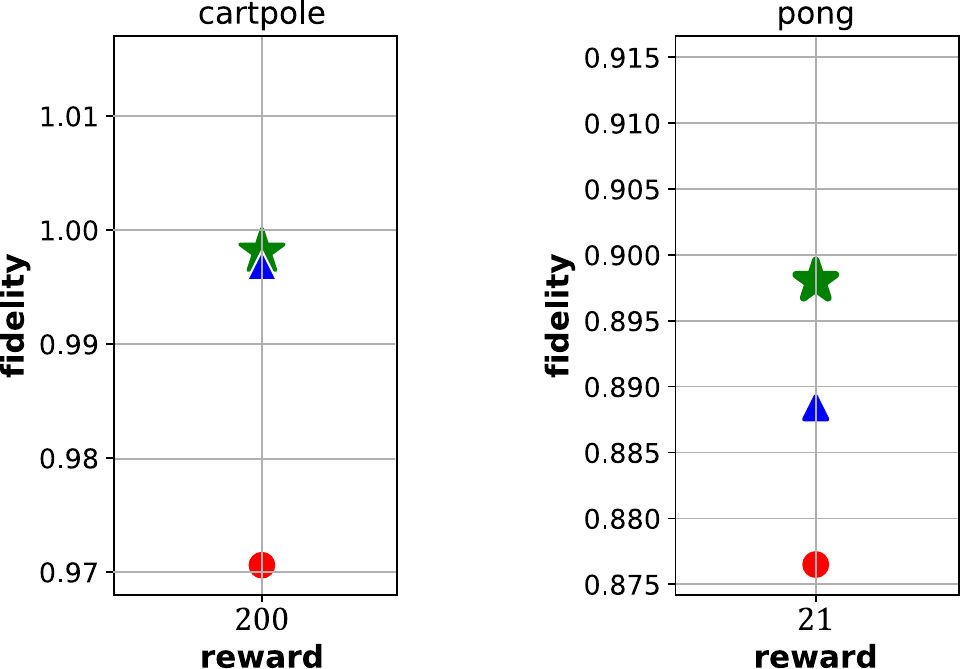}
\end{subfigure}
\begin{subfigure}[!t]{0.5\textwidth}
  \centering
  \includegraphics[scale=0.4]{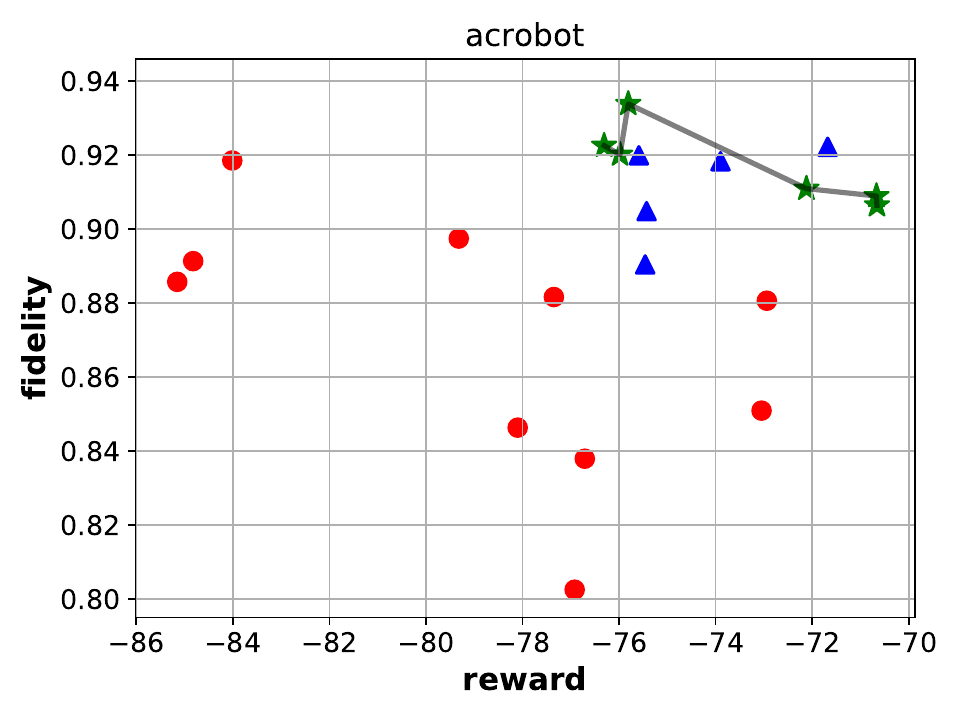}
\end{subfigure}
\begin{subfigure}[!t]{0.5\textwidth}
  \centering
  \includegraphics[scale=0.4]{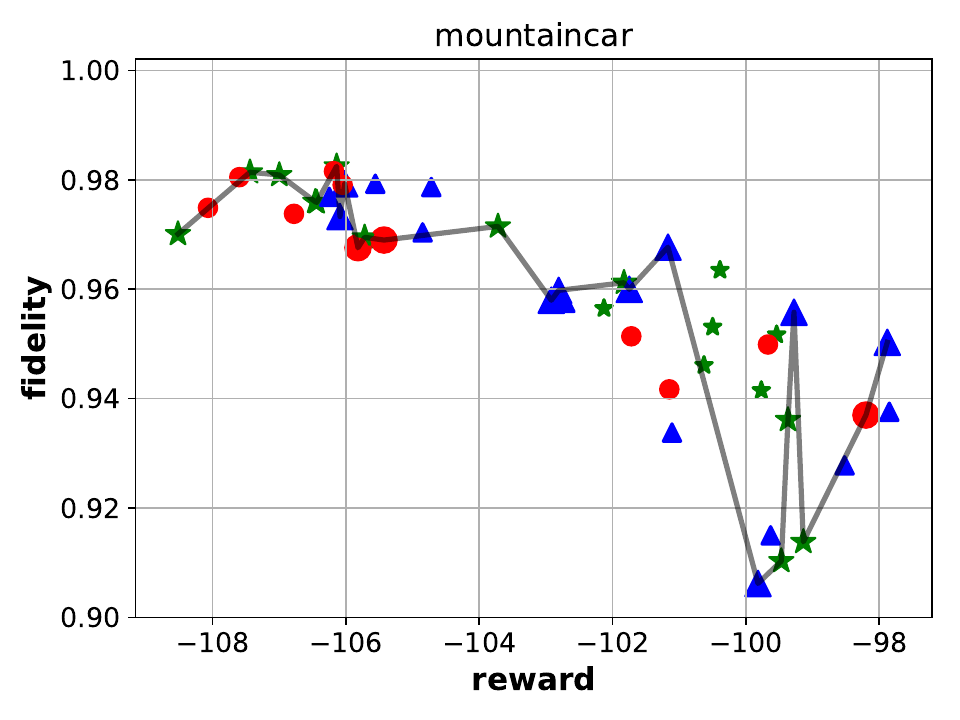}
\end{subfigure}
\begin{subfigure}[!t]{0.5\textwidth}
  \centering
  \includegraphics[scale=0.4]{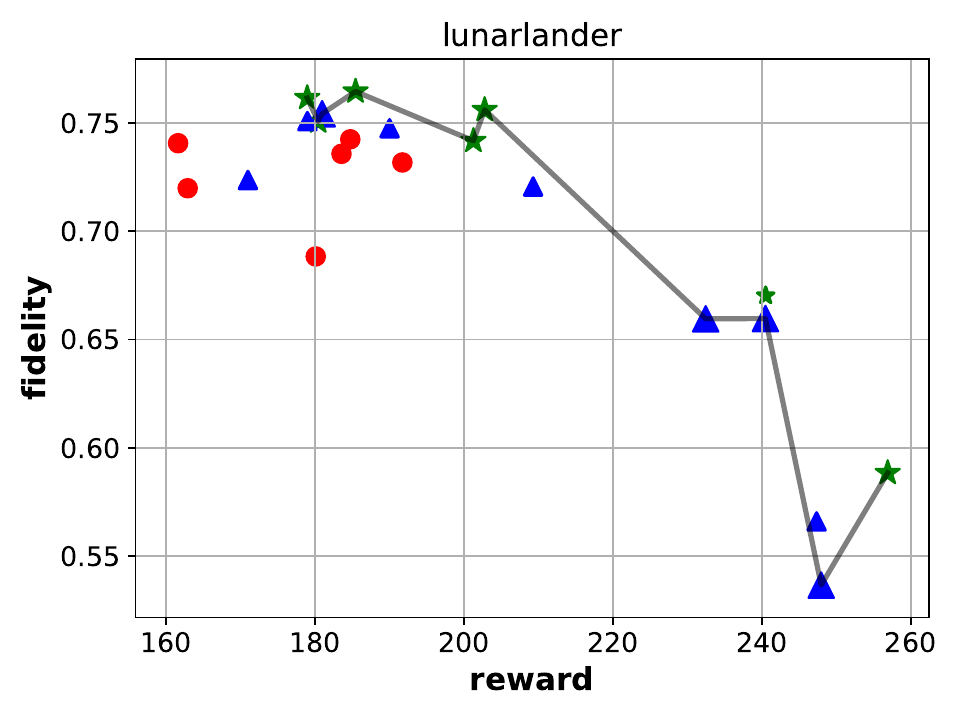}
\end{subfigure}
\begin{subfigure}[!t]{0.5\textwidth}
  \centering
  \includegraphics[scale=0.4]{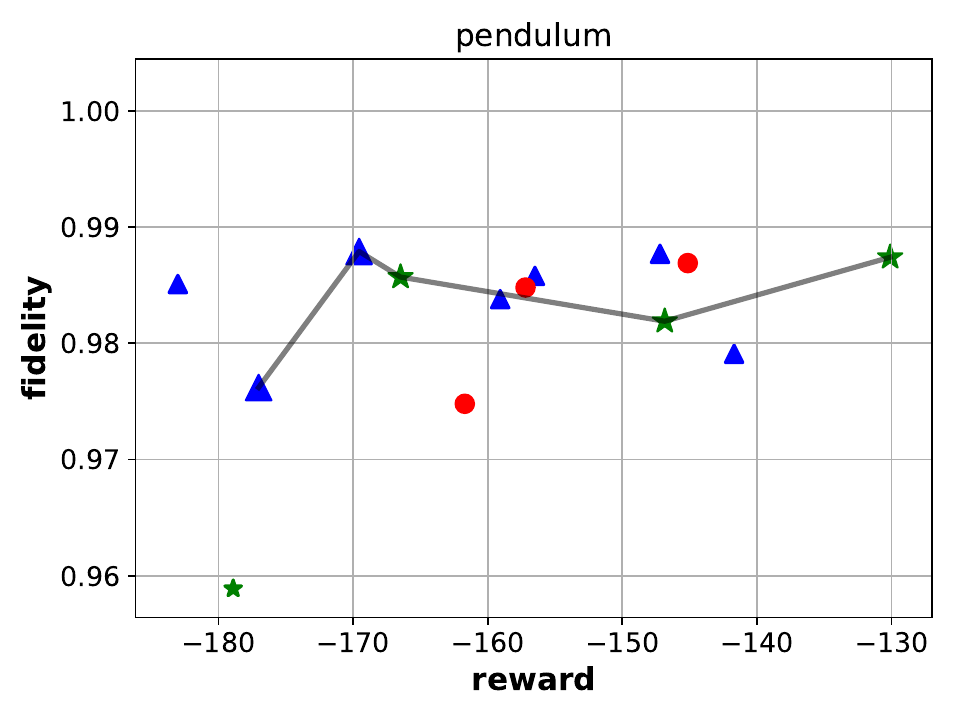}
\end{subfigure}
\begin{subfigure}[!t]{0.5\textwidth}
  \centering
  \includegraphics[scale=0.4]{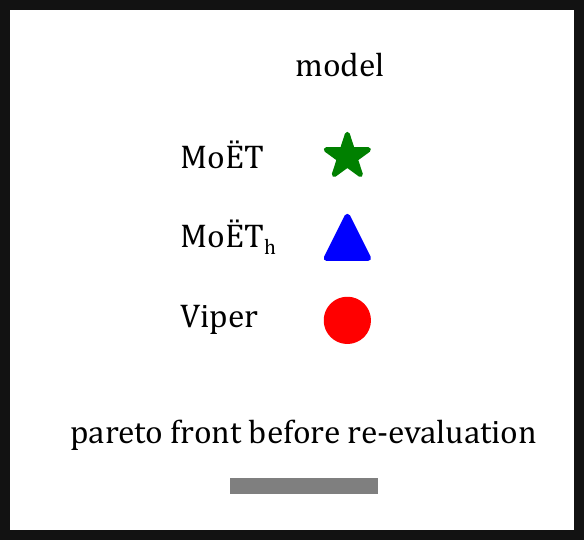}
\end{subfigure}
\caption{
  \textbf{Performance generalization of models}.
  Models on the Pareto fronts (Figure~\ref{fig:pre-evaluation}) are re-evaluated.
  Black solid line connects models that were on the global Pareto front before re-evaluation.
}
\label{fig:post-evaluation}
\end{figure*}

\textbf{Performance generalization of models}.
In the supervised learning setting, after the best models are selected based on their performance on a validation set, they are re-evaluated on a test set to get a better estimate of their performance on the new data.
In RL setting there is no direct analogy to validation and test datasets, but the models can be re-evaluated after the selection is performed.
After we identify the best models on the Pareto fronts (Figure~\ref{fig:pre-evaluation}), we re-evaluate their performance by running them again through the RL environment.
Figure~\ref{fig:post-evaluation} shows the achieved performance of these models after re-evaluation.
In the case of \Cartpole and \Pong performance before and after re-evaluation are very similar. %
In the case of \Acrobot, \Mountaincar and \Lunarlander, models that were on the global Pareto front are mostly still on the global Pareto front in the re-evaluation.
Moreover, \MOET and \MOETH models dominate \Viper models in most of the cases.
\Pendulum environment behaves more stochastically -- evaluating policy (done across $100$ episodes) can exhibit significantly different reward from evaluation to evaluation, making results more inconclusive.
However, all models achieve great fidelity level, and reward that is close to the \DRL agent one.
Considering high performance, differences in performance between models are minor.
Scores of the points that were on the global Pareto front are presented in a tabular form in \ref{sec:eval-appendix}.

Following the previous analysis, we conclude that \MOET and \MOETH models provide better performance (in terms of reward and fidelity) compared to \Viper in most of the cases,
demonstrating that \MOET is a valuable technique to be considered when looking for a verifiable RL policy.

\begin{figure}
  \includegraphics[scale=0.55,center]{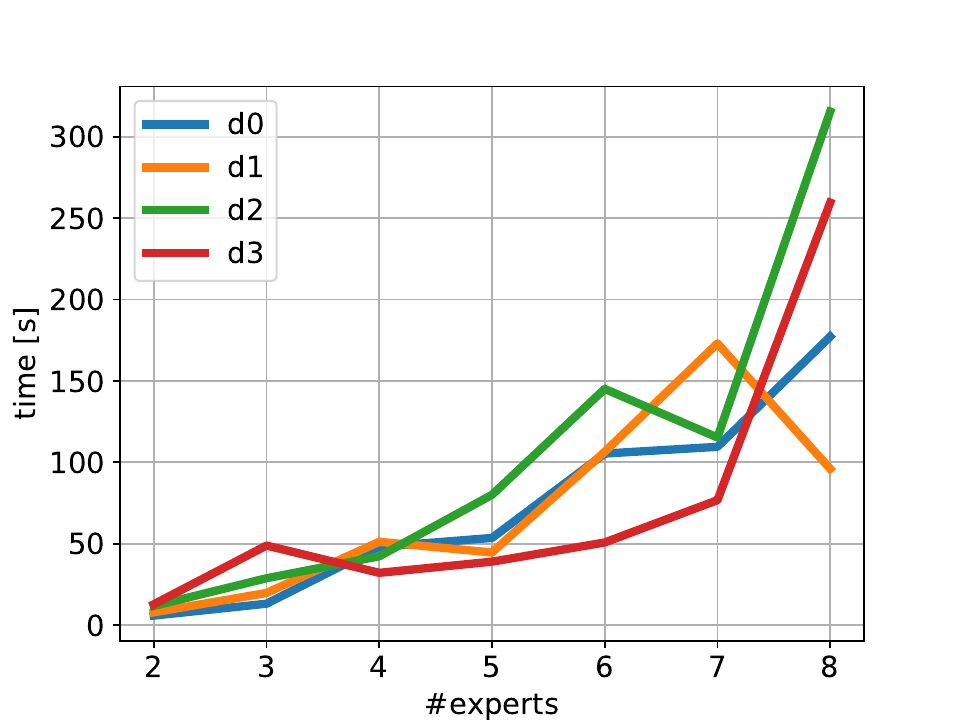}
  \caption{Verification times.}
  \label{fig:verification-times}
\end{figure}

\textbf{Verification}.
We perform verification of \MOETHard policies obtained in our experiments according to the procedure described in Section~\ref{sec:moet}.
All models considered in this experiment successfully pass the verification procedure.
To better understand the scalability of our verification procedure, we report the verification times needed to verify policies for different number of experts and expert depths in \fig~\ref{fig:verification-times}.
The verification times generally increase with the number of experts.
\MOETHard policies with 2 experts take from $5.5$s to $11.7$s for verification, while the verification times for 8 experts can go up to as much as $336$s.
This corresponds to the complexity of the logical formula obtained with an increase in the number of experts.
While the effect of expert depths on verification times is visible in a case of few experts,
with the increase of experts it is less noticeable,
thus indicating that the number of experts has more influence on the verification times than expert depths.
We run the verification on Intel i7-7600, 2.80GHz, 16 GB LPDDR3.
We show example SMT formula (of \Viper and \MOETH policies) in \ref{sec:smt:example}.

\begin{figure*}[!t]
\begin{subfigure}[!t]{1.\textwidth}
  \centering
  \includegraphics[scale=0.5]{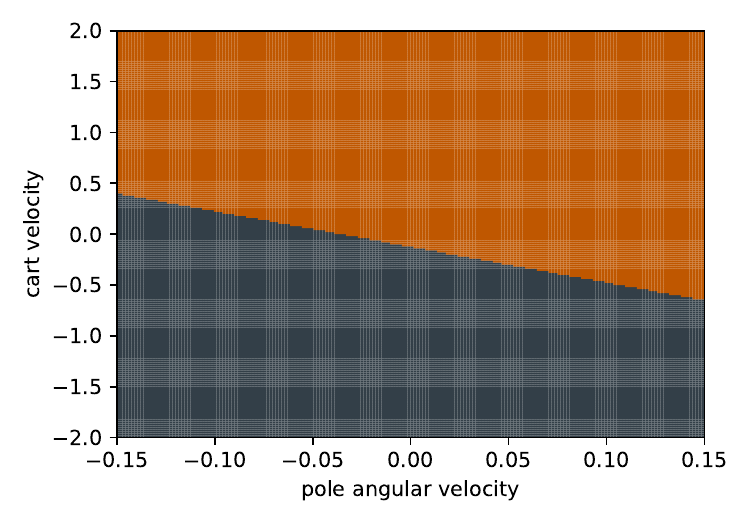}
\end{subfigure}
\begin{subfigure}[!t]{0.5\textwidth}
  \centering
  \includegraphics[scale=0.5]{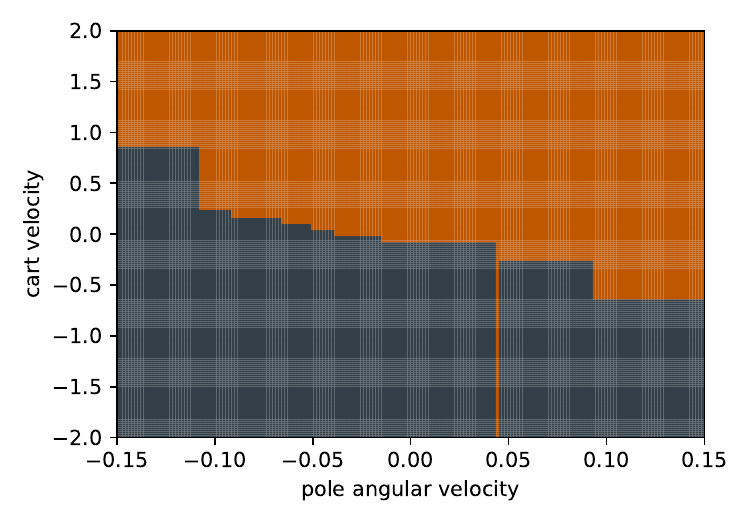}
\end{subfigure}
\begin{subfigure}[!t]{0.5\textwidth}
  \centering
  \includegraphics[scale=0.5]{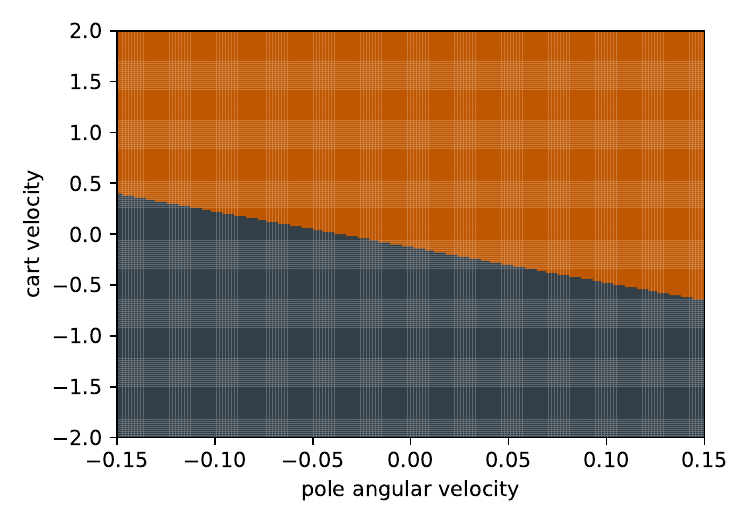}
\end{subfigure}
\caption{
  \textbf{Visualizing \DRL (top), \Viper (bottom left) and \MOETH (bottom right) policies on \Cartpole}.
  X-axis represents pole angular velocity and y-axis cart velocity, which are the most discriminatory features
  (topmost nodes in the \Viper decision tree policy).
  Other features, cart position and pole angle, are set to $0$ (center position with pole upright).
  Gray color represents points where agent takes action \textit{left}, and orange points when agent takes action \textit{right}.
}
\label{fig:cartpole_visualization}
\end{figure*}

\textbf{Expressiveness}.
We provide a simple qualitative comparison of best \Viper and \MOETH policies, by contrasting them to \DRL policy on a \Cartpole environment.
The figure \ref{fig:cartpole_visualization} visualizes these policies and demonstrates that
\MOETH policy much more closely resembles the \DRL policy thanks to its ability to represent hyperplanes of arbitrary orientation,
while \DT policy obtained by \Viper approximates \DRL policy by axis perpendicular hyperplanes.
The \MOETH policy presented is equivalent to the following program:
{{\verb|if| $2.18 * cp + 7.22 * cv + 20.64 * pa + 25.33 * pv > -1$ \verb|then| go right \verb|else| go left}},
where $cp$ and $cv$ are cart position and velocity,
and $pv$ and $pa$ pole angle and its angular velocity.

\textbf{Supervised learning.}
We evaluated the performance of \Tool and \MOETH in the supervised regime on three real-world datasets. 
Two datasets (German credit and Adult income) come from the UCI ML repository \cite{frank2011uci}, 
whereas the Fetal health dataset is a publicly available dataset that can be found on Kaggle. 
We summarize the properties of the datasets that we use in Table~\ref{tab:datasets}.

\begin{table}
	\caption{For each dataset used in the experimental evaluation we provide its name, the number of instances it contains (Size), numbers of instances per set after splitting the data into training, validation, and testing sets (Split) and total number of features (Features)}
	\centering
	\begin{tabular}{l|r|c|r}
          \toprule
          \TNC{1}{\HighlightCell{Dataset}} \vline & \TNC{1}{\HighlightCell{Size}} \vline & \TNC{1}{\HighlightCell{Split (train/test/val)}} \vline & \TNC{1}{\HighlightCell{Features}} \\
          \midrule
	  Adult income  & 48,842 & 34,189 / \hphantom{1}6,783 / \hphantom{1}6,784 & 14 \\ 
	  German credit & 1,000  & \hphantom{11,}700          / \hphantom{11,}150 / \hphantom{11,}150 & 10 \\
	  Fetal health  & 2,126  & \hphantom{1}1,488          / \hphantom{11,}319 / \hphantom{11,}319 & 21 \\
          \bottomrule
	\end{tabular}
	\label{tab:datasets}
\end{table}

In the \textit{Adult income} dataset \cite{kohavi1996scaling} the goal is to predict whether an income is greater than 50K dollars. 
In the \textit{German credit} dataset, the goal is to classify bank account holders into two classes -- good or bad. 
In the \textit{Fetal health} dataset, the goal is to predict whether a fetus is healthy or not based on the features extracted 
from cardiotocogram examination.

We compared \Tool with other supervised learning models which would require similar effort and tools to be verified: 
decision tree, support vector classifier (SVC) with linear kernel, ridge logistic regression and lasso logistic regression. 
The results are evaluated by F1 score and accuracy. The hyperparameters of compared models are tuned on validation set.
The results evaluated on test set with 95\% confidence intervals for \textit{Fetal health}, \textit{German credit}, and 
\textit{Adult income} datasets are presented in Tables~\ref{tab:fetal-health},~\ref{tab:credit}, and~\ref{tab:adult}, 
respectively. It can be observed 
that \Tool is the best performing model with exception of SVC being better on German credit data 
according to accuracy (but not F1 score). Therefore, it can be 
concluded that \Tool can also be successfully applied in the case of supervised learning problems.

\begin{table}
	\caption{Prediction performance of classifiers - \textit{Fetal health} dataset}
	\centering
	\begin{tabular}{c|c|c}
          \toprule
          \TNC{1}{\HighlightCell{model/metrics}} \vline & \TNC{1}{\HighlightCell{F1 score}} \vline & \TNC{1}{\HighlightCell{Accuracy}} \\
          \midrule
		Decision tree             & 0.852 $\pm$  0.004    & 0.939 $\pm$ 0.004    \\
		Lasso logistic regression & 0.797 $\pm$ 0.000    & 0.915 $\pm$ 0.000    \\
		\MOEHard                 & 0.880 $\pm$ 0.001     & 0.950 $\pm$ 0.001    \\
		\Tool                 & {\bf 0.891 $\pm$ 0.001}   & {\bf 0.955 $\pm$ 0.001}    \\
		Ridge logistic regression & 0.739 $\pm$ 0.000     & 0.903 $\pm$ 0.000    \\
		SVC                       & 0.762 $\pm$ 0.000     & 0.906 $\pm$ 0.000    \\
          \bottomrule
	\end{tabular}
	\label{tab:fetal-health}
\end{table}

\begin{table}
	\caption{Prediction performance of classifiers - \textit{German credit} dataset}
	\centering
	\begin{tabular}{c|c|c}
          \toprule
          \TNC{1}{\HighlightCell{model/metrics}} \vline & \TNC{1}{\HighlightCell{F1 score}} \vline & \TNC{1}{\HighlightCell{Accuracy}} \\
          \midrule
		Decision tree             & 0.759 $\pm$  0.000    & 0.637 $\pm$ 0.000    \\
		Lasso logistic regression & 0.797 $\pm$ 0.000     & 0.667 $\pm$ 0.000    \\
		\MOEHard                 & 0.759 $\pm$ 0.003     & 0.638 $\pm$ 0.004    \\
	        \Tool                    & {\bf 0.808 $\pm$ 0.003}     & 0.687 $\pm$ 0.004    \\
		Ridge logistic regression & 0.792 $\pm$ 0.000     & 0.660 $\pm$ 0.000    \\
		SVC                       & 0.799 $\pm$ 0.000     & 0.{\bf 693 $\pm$ 0.000}    \\
          \bottomrule
	\end{tabular}
	\label{tab:credit}
\end{table}

\begin{table}
	\caption{Prediction performance of classifiers - \textit{Adult income} dataset}
	\centering
	\begin{tabular}{c|c|c}
          \toprule
          \TNC{1}{\HighlightCell{model/metrics}} \vline & \TNC{1}{\HighlightCell{F1 score}} \vline & \TNC{1}{\HighlightCell{Accuracy}} \\
          \midrule
		Decision tree             & 0.661 $\pm$  0.003   & 0.852 $\pm$ 0.001    \\
		Lasso logistic regression & 0.536 $\pm$ 0.000    & 0.820 $\pm$ 0.000    \\
		\MOEHard                 & {\bf 0.676 $\pm$ 0.000}     & 0.854 $\pm$ 0.000    \\
		\Tool                 & {\bf 0.674 $\pm$ 0.004}    & {\bf 0.860 $\pm$ 0.001}    \\
		Ridge logistic regression & 0.529 $\pm$ 0.000     & 0.819 $\pm$ 0.000    \\
		SVC                       & 0.406 $\pm$ 0.000     & 0.805 $\pm$ 0.000    \\
          \bottomrule
	\end{tabular}
	\label{tab:adult}
\end{table}

\Section{Conclusion}
\label{sec:Conclusion}
We introduced \Tool, a technique based on \MOE with decision trees as experts and formulated a learning algorithm to train \Tool models.
To the best of our knowledge, this approach is the first to combine standard non-differentiable \DT experts with \MOE approach.
Furthermore, we used \MOET in RL setting by mimicking \DRL agents, in this way constructing RL policies that can be verified 
and are more interpretable than the \DRL agents themselves.
We showed a procedure to translate \MOET policies into SMT logic providing rich means for verification,
and showed that \MOET models perform better than the previous state-of-the-art approach \Viper and that they are also 
useful in the supervised regime.

\textbf{ACKNOWLEDGMENTS.} This work was supported by NSF grant CCF-1718903 to SK.

\bibliographystyle{elsarticle-num}
\bibliography{bib}

\newpage
\appendix

\Section{Viper Algorithm}
\label{sec:viper-algorithm}

\begin{algorithm*}[!t]
\caption{\Viper training~\cite{BastaniETAL18VerifiableRL}}\label{alg:viper}
\begin{algorithmic}[1]
\Procedure{\Viper(MDP $e$, Teacher $\pi_t$, Q-function $Q^{\pi_t}$, Iterations $N$)}{}
\State Initialize dataset and student: $D \gets \emptyset, \pi_{s_0} \gets \pi_t$
\For {$i \gets 1$ to $N$}
\State Sample trajectories and aggregate: $D \gets D \cup \{(s, \pi_t(s))\sim d^{\pi_{s_{i-1}}}(e)\}$
\State Sample dataset using Q values: $D_s \gets \{(s, a)\in I \sim D\}$
\State Train decision tree: $\pi_{s_i} \gets fit\_tree(D_s)$ \label{algline:fitting}
\EndFor
\State \Return Best policy $\pi_s \in \{\pi_{s_1}, ..., \pi_{s_N}\}$.
\EndProcedure
\end{algorithmic}
\end{algorithm*}

Viper algorithm is shown in Algorithm~\ref{alg:viper}.
\Section{Environments}
\label{sec:environments}

In this section we provide a brief description of environments we used in our experiments.
We used five environments from \OpenAIGym: \Cartpole, \Acrobot, \Mountaincar, \Lunarlander, \Pong and \Pendulum.

\subsection{\Cartpole}
This environment consists of a cart and a rigid pole hinged to the cart, based on the system presented by Barto et al.~\cite{BartoETAL83ControlProblemsLearning}.
At the beginning pole is upright, and the goal is to prevent it from falling over.
Cart is allowed to move horizontally within predefined bounds, and controller chooses to apply either \textit{left} or \textit{right} force to the cart.
State is defined with four variables: $x$ (cart position), $\dot{x}$ (cart velocity), $\theta$ (pole angle), and $\dot{\theta}$ (pole angular velocity).
Game is terminated when the absolute value of pole angle exceeds $12^{\circ}$, cart position is more than $2.4$ units away from the center, or after $200$ successful steps; whichever comes first.
In each step reward of $+1$ is given, and the game is considered solved when the average reward is over $195$ in over 100 consecutive trials.

\subsection{\Acrobot}
This environment is analogous to a gymnast swinging on a horizontal bar, and consists of a two links and two joins, where the joint between the links is actuated.
The environment is based on the system presented by Sutton~\cite{Sutton96GeneralizationInRL}.
Initially both links are pointing downwards, and the goal is to swing the end-point (feet) above the bar for at least the length of one link.
The state consists of six variables, four variables consisting of $\sin$ and $\cos$ values of the joint angles, and two variables for angular velocities of the joints.
The action is either applying \textit{negative}, \textit{neutral}, or \textit{positive} torque on the joint.
At each time step reward of $-1$ is received, and episode is terminated upon successful reaching the height, or after $200$ steps, whichever comes first.
\Acrobot is an unsolved environment in that there is no reward limit under which is considered solved, but the goal is to achieve high reward.

\subsection{\Mountaincar}
This environment consists of a car positioned between two hills, with a goal of reaching the hill in front of the car.
The environment is based on the system presented by Moore~\cite{Moore90RobotControl}.
Car can move in a one-dimensional track, but does not have enough power to reach the hill in one go, thus it needs to build momentum going back and forth to finally reach the hill.
Controller can choose \textit{left}, \textit{right} or \textit{neutral} action to apply left, right or no force to the car.
State is defined by two variables, describing car position and car velocity.
In each step reward of $-1$ is received, and episode is terminated upon reaching the hill, or after $200$ steps, whichever comes first.
The game is considered solved if average reward over $100$ consecutive trials is no less than $-110$.

\subsection{\Lunarlander}
This environment consists of a space ship and a landing pad, to which the ship should land.
Controller can choose when to turn on the left engine, right engine or the main engine, thus controlling the movement of the ship.
State is defined by:
$x$ and $y$ coordinates of the lander,
$v_x$ and $v_y$ velocities in the $x$ and $y$ direction,
$\theta$ angle of the lander,
$\alpha$ angular velocity,
and two boolean values indicating if left or right leg is touching the ground.
Episode finishes when lander crashes or comes to rest, after which it received appropriate reward.
Firing main engine is $-0.3$ points, and each leg contact is $10$ points.
The game is considered solved if achieved reward is at least $200$ points.

\subsection{\Pong}
This is a classical Atari game of table tennis with two players.
Minimum possible score is $-21$ and maximum is $21$.

\subsection{\Pendulum}
The environment consists of a pendulum, and the goal is to swing it up so it stays upright.
State is defined by:
$\theta$---angle of the pendulum,
and $\omega$---angular velocity of the pendulum.
Note that the OpenAI gym environment instead of the state feature $\theta$ contains two features: $x$ (which is equal to $cos(\theta)$) and $y$ (which is equal to $sin(\theta)$).
Action available is applying torque to the pendulum.
In OpenAI gym action can take any value in range $[-2,2]$.
We discretize action space into $3$ possible actions corresponding to torque of $-2$, $0$, or $2$.
In each step reward  obtained is equal to $-(\theta^2 + 0.1
cdot\omega^2 + 0.001\cdot torque^2)$.
Thus, the maximum reward that can be obtained in a step is $0$, which occurs when pendulum is upright, with zero velocity, and $0$ torque is applied to the pendulum.
Episode is of length $200$.

\Section{Model training parameters}
\label{sec:params}

\subsection{\DRL Agent Training}

In this section we present the architectures and hyperparameters used to train \DRL agents for different environments.

For \Cartpole, we use policy gradient model as used in \Viper.
While we use the same model, we had to retrain it from scratch as the trained \Viper agent was not available.
We use $1$ hidden layer with $8$ neurons.
We set discount factor to $0.99$, number of epochs to $1,000$ and batch size to $50$.

For \Pong, we use a \DQN network~\cite{MnihETAL15DRL} model that is already trained (the same as used in \Viper).
This model originates from the \OpenAI baselines~\cite{openAIBaselines}.

For \Acrobot, \Mountaincar and \Lunarlander, we implement our own version of dueling \DQN network following~\cite{wang2015dueling}.
We use $3$ hidden layers with $15$ neurons in each layer for \Mountaincar, and $50$ neurons in each layer for \Acrobot and \Lunarlander.
We set the learning rate to $0.001$,
batch size to $30$ in \Mountaincar, $50$ in \Acrobot and \Lunarlander,
step size to $10,000$ and number of epochs to $80,000$ in \Mountaincar, $50,000$ in \Acrobot and \Lunarlander.
We checkpoint a model every $5,000$ steps and pick the best performing one in terms of achieved reward.

\subsection{\Viper and \Tool Training}
We used $40$ iterations of \Dagger, and $200,000$ as a maximum number of samples for training student policies.
During evaluation, cumulative reward is averaged across $100$ runs in a given environment ($250$ in a case of \Cartpole).

We trained \Viper for varying value of the tree maximum depth.
The values used are:
$[1,15]$ in \Cartpole, $[1,20]$ in \Acrobot, $[1,20]$ in \Mountaincar, $[1,30]$ In \Lunarlander, and $[1,35]$ in \Pong.

We trained \MOET models for varying number of experts and their maximum depths.
The number of experts used are:
$[2, 8]$ in \Cartpole, $[2, 8] \cup [15,16]$ in \Acrobot, $[2, 8] \cup \{12, 16\}$ in \Mountaincar, $[2, 8]$ in \Lunarlander, and $\{2, 4, 8, 16, 32\}$ in \Pong.
The maximum depths of experts are:
$[0,7]$ in \Cartpole, $[0,15]$ in \Acrobot, $[0,11]$ in \Mountaincar, $[0,20]$ in \Lunarlander, and $[0,29]$ in \Pong.
We used following learning rates for training \Tool models: $\{1, 0.3, 0.1, 0.01, 0.001, 0.0001, 0.00001\}$,
while for the learning rate decay we used $1$ (no decay) and $0.97$ (learning rate is multiplied by this value after each epoch).
As for the maximum number of epochs for \Tool training procedure we used values: $\{50, 100, 500\}$.

\subsection{Compute}

To run our experiments we used a cluster with nodes of the following configuration:
Xeon CPU E5-2650 v3 (Haswell): 10 cores per socket (20 cores/node),
2.30GHz, 128 GB DDR4-2133. We used up to 10 such nodes when scheduling
our experiments.
\Section{SMT translation example}
\label{sec:smt:example}

\begin{figure*}[t]
  \centering
  \includegraphics[scale=0.5]{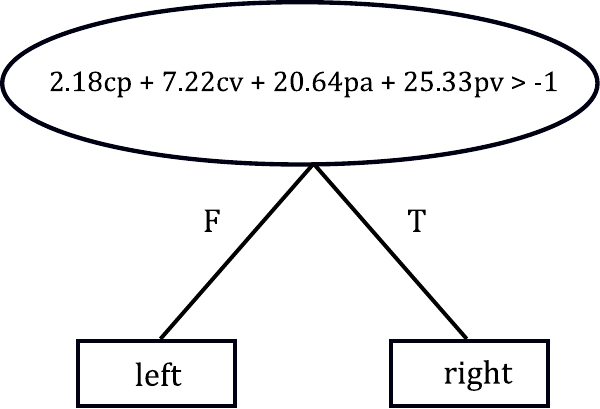}
  \caption{
    Example \Cartpole \MOETH policy.
  }
  \label{fig:carpole:moeth:policy}
\end{figure*}

The \Cartpole \MOETH policy presented in \fig~\ref{fig:cartpole_visualization} is shown in \fig~\ref{fig:carpole:moeth:policy}.
SMT formula that would encode the policy part (mapping input to a model decision) of \Cartpole verification formula would look as follows:
\CodeIn{If(2.18cp + 7.22cv + 20.64pa + 25.33pv > -1, 1, 0)}.
This \MOETH policy consists of the gating expressed by the inequality and two trivial expert decision trees of depth $0$.
Therefore, second and third part of the \CodeIn{If} formula are trivial.
In case that decision trees were nontrivial, those parts of the formula would be expanded with nested if expressions.

A simple depth $2$ \Viper policy for \Cartpole is shown in \fig~\ref{fig:carpole:moeth:policy}.
SMT formula that would encode the policy part of this formula would look like following:
\CodeIn{If(pv < -0.033, If(pa < 0.039, 0, 1), If(pa < -0.037, 0, 1))}

The full formula for \Cartpole environment verification contains additional details, it is 
the conjunction of the formula encoding the policy, the safety requirements and the environment dynamics,
as illustrated by the formula in Section~\ref{sec:moet}.

\begin{figure*}[t]
  \centering
  \includegraphics[scale=0.5]{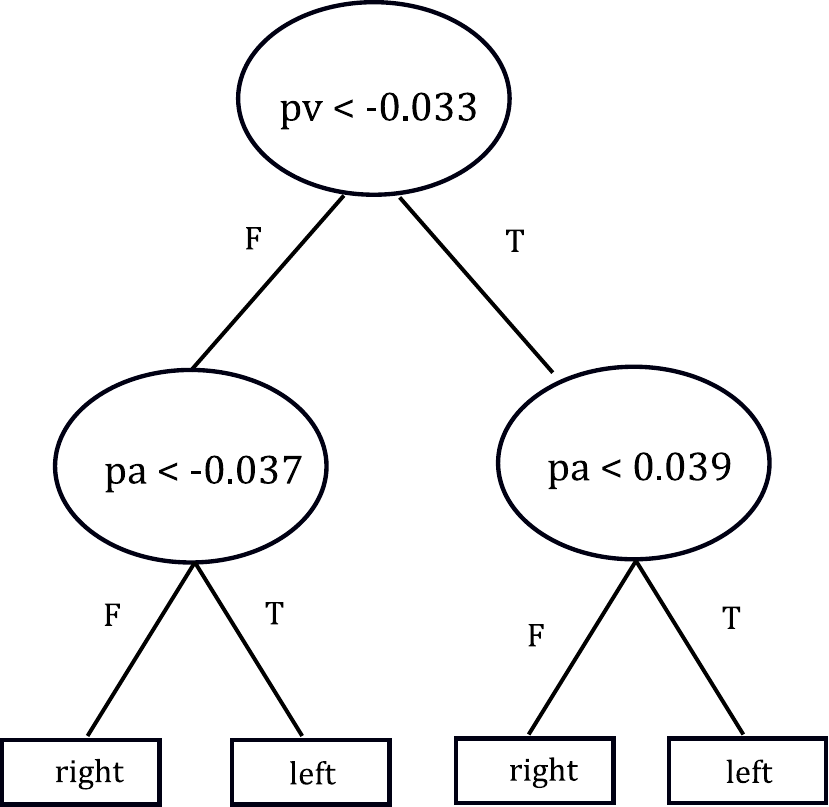}
  \caption{
    Example \Cartpole \Viper policy.
  }
\end{figure*}
\Section{Evaluation Results}
\label{sec:eval-appendix}

Tables~\ref{tbl:cartpole:global-Pareto},~\ref{tbl:acrobot:global-Pareto},~\ref{tbl:mountaincar:global-Pareto},~\ref{tbl:lunarlander:global-Pareto},~\ref{tbl:pong:global-Pareto},~\ref{tbl:pendulum:global-Pareto} show data about models on the global Pareto front presented in \fig~\ref{fig:pre-evaluation} of Section~\ref{sec:Experiments}.

Tables~\ref{tbl:cartpole:reeval-Pareto},~\ref{tbl:acrobot:reeval-Pareto},~\ref{tbl:mountaincar:reeval-Pareto},~\ref{tbl:lunarlander:reeval-Pareto},~\ref{tbl:pong:reeval-Pareto},~\ref{tbl:pendulum:reeval-Pareto} show data about the models on the global Pareto after reevaluation is performed. This corresponds to data presented in \fig~\ref{fig:post-evaluation} of Section~\ref{sec:Experiments}.

\begin{table}[H]
  \caption{\Cartpole: global Pareto front data}
  \label{tbl:cartpole:global-Pareto}
  \footnotesize
  \centering
  \begin{tabular}{cccc}
    \toprule
    \TNC{1}{\HighlightCell{Model}} & \TNC{1}{\HighlightCell{Configuration}} & \TNC{1}{\HighlightCell{Reward}} & \TNC{1}{\HighlightCell{Fidelity}} \\
    \midrule
    \MOET & E2-D0 & $200.00$ & $0.998$ \\
    \bottomrule
  \end{tabular}
\end{table}

\begin{table}[H]
  \caption{\Acrobot: global Pareto front data}
  \label{tbl:acrobot:global-Pareto}
  \footnotesize
  \centering
  \begin{tabular}{cccc}
    \toprule
    \TNC{1}{\HighlightCell{Model}} & \TNC{1}{\HighlightCell{Configuration}} & \TNC{1}{\HighlightCell{Reward}} & \TNC{1}{\HighlightCell{Fidelity}} \\
    \midrule
    \MOET & E16-D11 & $-72.12$ & $0.936$ \\
    \MOET & E15-D11 & $-71.95$ & $0.936$ \\
    \MOET & E15-D11 & $-71.81$ & $0.921$ \\
    \MOET & E16-D9 & $-71.67$ & $0.921$ \\
    \MOET & E16-D0 & $-69.83$ & $0.916$ \\
    \MOET & E16-D0 & $-68.68$ & $0.907$ \\
    \bottomrule
  \end{tabular}
\end{table}

\begin{table}[H]
  \caption{\Mountaincar: global Pareto front data}
  \label{tbl:mountaincar:global-Pareto}
  \footnotesize
  \centering
  \begin{tabular}{cccc}
    \toprule
    \TNC{1}{\HighlightCell{Model}} & \TNC{1}{\HighlightCell{Configuration}} & \TNC{1}{\HighlightCell{Reward}} & \TNC{1}{\HighlightCell{Fidelity}} \\
    \midrule
    \MOETH & E6-D9 & $-107.00$ & $0.984$ \\
    \MOET & E6-D7 & $-106.83$ & $0.984$ \\
    \MOET & E16-D7 & $-105.90$ & $0.983$ \\
    \MOET & E7-D8 & $-104.28$ & $0.982$ \\
    \MOET & E3-D7 & $-103.86$ & $0.979$ \\
    \MOET & E3-D10 & $-103.82$ & $0.977$ \\
    \MOETH & E3-D6 & $-103.77$ & $0.977$ \\
    \MOET & E7-D5 & $-103.75$ & $0.974$ \\
    \MOET & E3-D7 & $-103.22$ & $0.973$ \\
    \Viper & D12 & $-102.83$ & $0.973$ \\
    \MOET & E2-D8 & $-102.45$ & $0.972$ \\
    \Viper & D11 & $-102.05$ & $0.972$ \\
    \MOETH & E4-D4 & $-101.40$ & $0.971$ \\
    \MOET & E5-D5 & $-101.09$ & $0.966$ \\
    \MOETH & E8-D5 & $-100.97$ & $0.962$ \\
    \MOETH & E4-D5 & $-100.96$ & $0.961$ \\
    \MOETH & E2-D8 & $-100.95$ & $0.961$ \\
    \MOETH & E4-D5 & $-98.85$ & $0.960$ \\
    \MOETH & E4-D5 & $-98.70$ & $0.950$ \\
    \MOET & E4-D4 & $-97.84$ & $0.943$ \\
    \Viper & D5 & $-97.46$ & $0.938$ \\
    \MOET & E7-D2 & $-97.39$ & $0.922$ \\
    \MOET & E4-D2 & $-96.96$ & $0.914$ \\
    \MOETH & E6-D1 & $-96.78$ & $0.912$ \\
    \bottomrule
  \end{tabular}
\end{table}

\begin{table}[H]
  \caption{\Lunarlander: global Pareto front data}
  \label{tbl:lunarlander:global-Pareto}
  \footnotesize
  \centering
  \begin{tabular}{cccc}
    \toprule
    \TNC{1}{\HighlightCell{Model}} & \TNC{1}{\HighlightCell{Configuration}} & \TNC{1}{\HighlightCell{Reward}} & \TNC{1}{\HighlightCell{Fidelity}} \\
    \midrule
    \MOET & E8-D17 & $204.13$ & $0.792$ \\
    \MOET & E7-D17 & $210.79$ & $0.767$ \\
    \MOET & E8-D17 & $217.33$ & $0.765$ \\
    \MOET & E8-D17 & $225.24$ & $0.755$ \\
    \MOETH & E8-D17 & $229.20$ & $0.747$ \\
    \MOET & E6-D17 & $230.67$ & $0.743$ \\
    \MOETH & E7-D0 & $239.96$ & $0.666$ \\
    \MOETH & E7-D0 & $241.25$ & $0.635$ \\
    \MOET & E6-D3 & $253.64$ & $0.628$ \\
    \MOETH & E7-D0 & $261.86$ & $0.547$ \\
    \bottomrule
  \end{tabular}
\end{table}

\begin{table}[H]
  \caption{\Pong: global Pareto front data}
  \label{tbl:pong:global-Pareto}
  \footnotesize
  \centering
  \begin{tabular}{cccc}
    \toprule
    \TNC{1}{\HighlightCell{Model}} & \TNC{1}{\HighlightCell{Configuration}} & \TNC{1}{\HighlightCell{Reward}} & \TNC{1}{\HighlightCell{Fidelity}} \\
    \midrule
    \MOET & E16-D21 & $21.00$ & $0.896$ \\
    \bottomrule
  \end{tabular}
\end{table}

\begin{table}[H]
  \caption{\Pendulum: global Pareto front data}
  \label{tbl:pendulum:global-Pareto}
  \footnotesize
  \centering
  \begin{tabular}{cccc}
    \toprule
    \TNC{1}{\HighlightCell{Model}} & \TNC{1}{\HighlightCell{Configuration}} & \TNC{1}{\HighlightCell{Reward}} & \TNC{1}{\HighlightCell{Fidelity}} \\
    \midrule
    \MOET & E8-D16 & $-170.00$ & $0.988$ \\
    \MOETH & E7-D17 & $-141.17$ & $0.988$ \\
    \MOET & E4-D15 & $-134.06$ & $0.988$ \\
    \MOET & E6-D13 & $-127.25$ & $0.985$ \\
    \MOETH & E2-D12 & $-120.31$ & $0.979$ \\
    \bottomrule
  \end{tabular}
\end{table}

\begin{table}[H]
  \caption{\Cartpole: reevaluation Pareto}
  \label{tbl:cartpole:reeval-Pareto}
  \footnotesize
  \centering
  \begin{tabular}{cccc}
    \toprule
    \TNC{1}{\HighlightCell{Model}} & \TNC{1}{\HighlightCell{Configuration}} & \TNC{1}{\HighlightCell{Reward}} & \TNC{1}{\HighlightCell{Fidelity}} \\
    \midrule
    \MOET & E2-D0 & $200.00$ & $0.998$ \\
    \bottomrule
  \end{tabular}
\end{table}

\begin{table}[H]
  \caption{\Acrobot: reevaluation Pareto}
  \label{tbl:acrobot:reeval-Pareto}
  \footnotesize
  \centering
  \begin{tabular}{cccc}
    \toprule
    \TNC{1}{\HighlightCell{Model}} & \TNC{1}{\HighlightCell{Configuration}} & \TNC{1}{\HighlightCell{Reward}} & \TNC{1}{\HighlightCell{Fidelity}} \\
    \midrule
    \MOET & E15-D11 & $-76.31$ & $0.923$ \\
    \MOET & E15-D11 & $-75.98$ & $0.920$ \\
    \MOET & E16-D11 & $-75.81$ & $0.934$ \\
    \MOET & E16-D9 & $-72.12$ & $0.911$ \\
    \MOET & E16-D0 & $-70.67$ & $0.909$ \\
    \MOET & E16-D0 & $-70.66$ & $0.907$ \\
    \bottomrule
  \end{tabular}
\end{table}

\begin{table}[H]
  \caption{\Mountaincar: reevaluation Pareto}
  \label{tbl:mountaincar:reeval-Pareto}
  \footnotesize
  \centering
  \begin{tabular}{cccc}
    \toprule
    \TNC{1}{\HighlightCell{Model}} & \TNC{1}{\HighlightCell{Configuration}} & \TNC{1}{\HighlightCell{Reward}} & \TNC{1}{\HighlightCell{Fidelity}} \\
    \midrule
    \MOET & E3-D7 & $-108.52$ & $0.970$ \\
    \MOET & E7-D8 & $-107.44$ & $0.981$ \\
    \MOET & E16-D7 & $-107.00$ & $0.981$ \\
    \MOET & E3-D7 & $-106.46$ & $0.976$ \\
    \MOET & E3-D10 & $-106.44$ & $0.976$ \\
    \MOET & E6-D7 & $-106.14$ & $0.983$ \\
    \MOETH & E3-D6 & $-106.09$ & $0.973$ \\
    \MOETH & E6-D9 & $-106.02$ & $0.979$ \\
    \Viper & D11 & $-105.82$ & $0.968$ \\
    \MOET & E2-D8 & $-105.72$ & $0.970$ \\
    \Viper & D12 & $-105.43$ & $0.969$ \\
    \MOET & E7-D5 & $-103.72$ & $0.972$ \\
    \MOETH & E8-D5 & $-102.92$ & $0.958$ \\
    \MOETH & E2-D8 & $-102.81$ & $0.960$ \\
    \MOET & E5-D5 & $-101.83$ & $0.961$ \\
    \MOETH & E4-D5 & $-101.75$ & $0.960$ \\
    \MOETH & E4-D4 & $-101.17$ & $0.968$ \\
    \MOETH & E6-D1 & $-99.82$ & $0.906$ \\
    \MOET & E4-D2 & $-99.47$ & $0.910$ \\
    \MOET & E4-D4 & $-99.37$ & $0.936$ \\
    \MOETH & E4-D5 & $-99.28$ & $0.956$ \\
    \MOET & E7-D2 & $-99.14$ & $0.914$ \\
    \Viper & D5 & $-98.20$ & $0.937$ \\
    \MOETH & E4-D5 & $-97.88$ & $0.950$ \\
    \bottomrule
  \end{tabular}
\end{table}

\begin{table}[H]
  \caption{\Lunarlander: reevaluation Pareto}
  \label{tbl:lunarlander:reeval-Pareto}
  \footnotesize
  \centering
  \begin{tabular}{cccc}
    \toprule
    \TNC{1}{\HighlightCell{Model}} & \TNC{1}{\HighlightCell{Configuration}} & \TNC{1}{\HighlightCell{Reward}} & \TNC{1}{\HighlightCell{Fidelity}} \\
    \midrule
    \MOET & E8-D17 & $178.93$ & $0.762$ \\
    \MOET & E6-D17 & $180.40$ & $0.751$ \\
    \MOETH & E8-D17 & $180.93$ & $0.754$ \\
    \MOET & E8-D17 & $185.42$ & $0.765$ \\
    \MOET & E7-D17 & $201.25$ & $0.742$ \\
    \MOET & E8-D17 & $202.76$ & $0.756$ \\
    \MOETH & E7-D0 & $232.45$ & $0.660$ \\
    \MOETH & E7-D0 & $240.48$ & $0.660$ \\
    \MOETH & E7-D0 & $247.97$ & $0.537$ \\
    \MOET & E6-D3 & $256.90$ & $0.588$ \\
    \bottomrule
  \end{tabular}
\end{table}

\begin{table}[H]
  \caption{\Pong: reevaluation Pareto}
  \label{tbl:pong:reeval-Pareto}
  \footnotesize
  \centering
  \begin{tabular}{cccc}
    \toprule
    \TNC{1}{\HighlightCell{Model}} & \TNC{1}{\HighlightCell{Configuration}} & \TNC{1}{\HighlightCell{Reward}} & \TNC{1}{\HighlightCell{Fidelity}} \\
    \midrule
    \MOET & E16-D21 & $21.00$ & $0.898$ \\
    \bottomrule
  \end{tabular}
\end{table}

\begin{table}[H]
  \caption{\Pendulum: reevaluation Pareto}
  \label{tbl:pendulum:reeval-Pareto}
  \footnotesize
  \centering
  \begin{tabular}{cccc}
    \toprule
    \TNC{1}{\HighlightCell{Model}} & \TNC{1}{\HighlightCell{Configuration}} & \TNC{1}{\HighlightCell{Reward}} & \TNC{1}{\HighlightCell{Fidelity}} \\
    \midrule
    \MOETH & E2-D12 & $-177.01$ & $0.976$ \\
    \MOETH & E7-D17 & $-169.55$ & $0.988$ \\
    \MOET & E4-D15 & $-166.47$ & $0.986$ \\
    \MOET & E6-D13 & $-146.85$ & $0.982$ \\
    \MOET & E8-D16 & $-130.11$ & $0.987$ \\
    \bottomrule
  \end{tabular}
\end{table}
  
\end{document}